\documentclass[letterpaper]{article} 
\usepackage[]{aaai25}  
\usepackage{times}  
\usepackage{helvet}  
\usepackage{courier}  
\usepackage[hyphens]{url}  
\usepackage{natbib}  
\usepackage{caption} 
\usepackage{latexsym}
\usepackage{comment}
\usepackage{float}
\usepackage{inconsolata}
\usepackage{scalerel,graphicx,xparse}
\usepackage{dsfont}
\usepackage{amsmath}
\usepackage{booktabs}
\usepackage{colortbl}

\usepackage{multirow}
\usepackage[table,xcdraw]{xcolor}
\usepackage{caption}
\usepackage{subcaption}

\newcommand{\task}{Text-to-SQL}
\newcommand{\ehrsql}{EHRSQL}

\newcommand{\llama}{Llama}

\title{Confidence Estimation for Error Detection in Text-to-SQL Systems}
\author {
    Oleg Somov\textsuperscript{\rm 1,2} and
    Elena Tutubalina\textsuperscript{\rm 1,3,4}
}
\affiliations {
    \textsuperscript{\rm 1}AIRI, Moscow, Russia \\
    \textsuperscript{\rm 2}MIPT, Dolgoprudny, Russia \\
    \textsuperscript{\rm 3}Sber AI, Moscow, Russia \\
    \textsuperscript{\rm 4}ISP RAS Research Center for Trusted Artificial Intelligence, Moscow, Russia \\
    somov@airi.net, tutubalina@airi.net
}

\begin{document}

\maketitle

\begin{abstract}
Text-to-SQL enables users to interact with databases through natural language, simplifying the retrieval and synthesis of information.
Despite the success of large language models (LLMs) in converting natural language questions into SQL queries, their broader adoption is limited by two main challenges: achieving robust generalization across diverse queries and ensuring interpretative confidence in their predictions. 
To tackle these issues, our research investigates the integration of selective classifiers into Text-to-SQL systems. 
We analyse the trade-off between coverage and risk using entropy based confidence estimation with selective classifiers and assess its impact on the overall performance of Text-to-SQL models. 
Additionally, we explore the models' initial calibration and improve it with calibration techniques for better model alignment between confidence and accuracy. 
Our experimental results show that encoder-decoder T5 is better calibrated than in-context-learning GPT 4 and decoder-only \llama{} 3, thus the designated external entropy-based selective classifier has better performance. 
The study also reveal that, in terms of error detection, selective classifier with a higher probability detects errors associated with irrelevant questions rather than incorrect query generations.
\end{abstract}

%
\begin{links}
    \link{Code}{https://github.com/runnerup96/error-detection-in-text2sql}
\end{links}

\section{Introduction}
\begin{figure}[t!]
  \centering    
  {\includegraphics[width=0.40\textwidth]{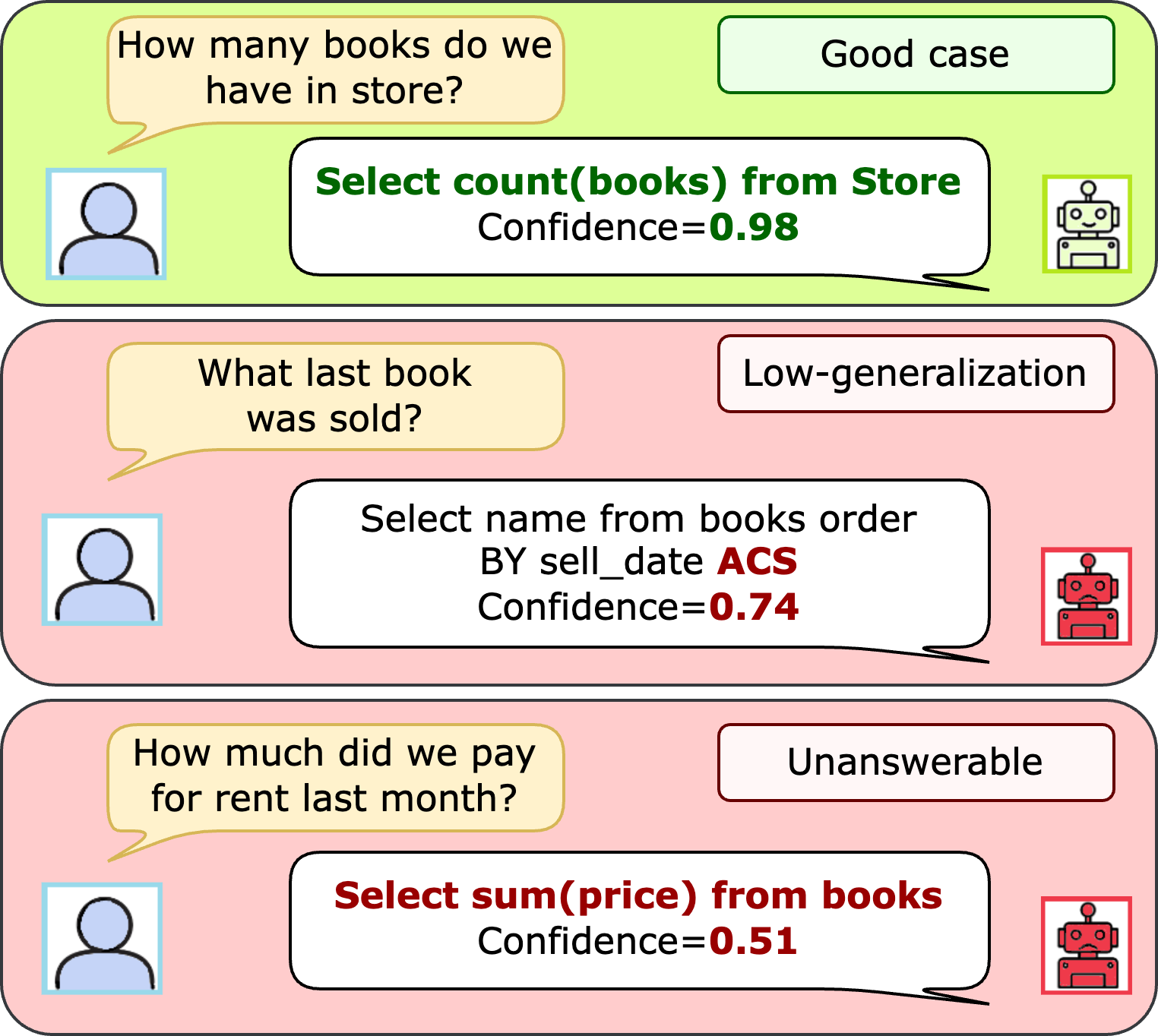}}
  \caption{The interaction scenario with \task{} system. There are three major scenarios where confidence is crucial - good generation detection, error generation, and unanswerable query detection.}
  \label{fig:interaction_scenario}
\end{figure}

Text-to-SQL parsing \cite{zelle1996learning,10.5555/3020336.3020416} aims at converting a natural language (NL) question to its corresponding structured query language (SQL) in the context of a relational database (schema). To effectively utilize Text-to-SQL models, users must clearly understand the model's capabilities, particularly the range of questions it can accurately respond to.  In this context, \textit{generalization ability} is crucial for ensuring accurate SQL query generation, while \textit{interpretative trustworthiness} is essential for minimizing false positives—instances where the model generates incorrect SQL queries that might be mistakenly perceived as correct.

Our work explores how model uncertainty estimates can be leveraged to detect erroneous predictions (Fig. \ref{fig:interaction_scenario}). We focus on scenarios where models struggle to generalize, such as low compositional generalization, where the model cannot form novel structures not seen in the training set, or low domain generalization, where the model fails to adapt to new schema elements or novel databases. We categorize these errors as \textit{low-generalization}. Additionally, we address cases where questions are unanswerable by the underlying database or require external knowledge, which we refer to as \textit{unanswerable}. Typically, models are trained on question-SQL pairs, so encountering such samples is considered out-of-distribution (OOD), and the model should ideally avoid generating a query.  Both types of errors result in incorrect responses, whether as non-executable SQL queries or executable queries that return false responses.

To study compositional and domain generalization in \task{}, several benchmarks and datasets have been developed over the years to better approximate real-world scenarios, address various aspects of model performance: complex queries involving join statements across multiple tables \cite{yu2018spider}, new and unseen database schemas \cite{gan2021exploring,lee2021kaggledbqa}, compositional train and test splits \cite{shaw-etal-2021-compositional,finegan-dollak-etal-2018-improving}, robustness test sets \cite{bakshandaeva-etal-2022-pauq,changdr}, dirty schema values and external knowledge requirements \cite{li2024can,wretblad2024understanding}, domain-specific datasets that feature unanswerable questions \cite{lee2022ehrsql}. To sum up, most prior work evaluates \textit{either} different types of generalization, noisy data \textit{or} model uncertainties. 


In this paper, we ask: \textit{can we identify error generations in \task{} LMs under distribution shift using uncertainty estimation}? Specifically, we examine this from the point of view of \textit{error detection} and \textit{calibration}, examining if the models' probability estimates align accurately with the actual correctness of the answers. We apply T5 \cite{raffel2020exploring}, GPT 4 \cite{achiam2023gpt}, and \llama{} 3 \cite{meta2024introducing} with a reject option\footnote{Selective classification, or classification with a reject option, attempts to abstain on examples that the model is likely to get wrong. This concept was introduced decades ago \cite{chow-etal-1957,chow1970optimum} for decision-making in scenarios where mistakes are costly but abstentions are allowed. In general, by allowing a classifier to abstain, one can improve the performance of a model at the cost of reducing coverage and classifying fewer samples. Further, we use the term ``selective Text-to-SQL'' or ``Text-to-SQL with a reject option''.} over popular SPIDER \cite{yu2018spider} and EHRSQL \cite{lee2022ehrsql}, covering general-domain and clinical domains. 
Our findings indicate that in distribution shift settings (cross-database or compositional), the selective classifier achieves high recall but low precision in error detection, leading to loss of total generated queries and deterioration in overall Text-to-SQL system quality. Our analysis also revealed that unanswerable queries are more likely to be detected using our confidence estimates than incorrect queries (Sec. \ref{sec:cs1}). Conversely, in a benchmark with unanswerable queries without a significant compositional or domain shift, the Text-to-SQL system with a selective classifier performs better overall in error detection with a lower rejection loss of correct queries.
Furthermore, we examined the calibration characteristics of logit-based confidence estimates (Sec. \ref{sec:cs2}). Under distribution shift, all fine-tuned models lacked proper calibration. Post-hoc calibration methods such as Platt Calibration and Isotonic Regression improved the initial models' calibration, underscoring the importance of calibration techniques in enhancing the reliability of model predictions. As a result, experiments demonstrate that while decoder-only models perform better on certain datasets (compositional or cross-database), encoder-decoder methods exhibit superior calibration for \task{} after post-hoc calibration. In Sec.~\ref{sec:cs3}, we did an in-depth analysis of the relation of selective classifier confidence and generated query complexity.

\section{Related work}
\paragraph{Uncertainty estimation and error detection}

The reliability of Text-to-SQL models or question answering systems, in general, is closely tied to their calibration ability for error detection and result interpretation.  Selective prediction,  where a model can choose to predict or abstain, has been a longstanding topic in machine learning \cite{chow-etal-1957, JMLR:v11:el-yaniv10a, dong-etal-2018-confidence}.

The rise of LLMs has highlighted the issue of hallucinations in NLP. Uncertainty estimation is a key research area for developing calibrated Text-to-SQL systems and reliable selective prediction algorithms.
Several recent works \cite{malinin2021uncertaintyestimationautoregressivestructured, van-der-poel-etal-2022-mutual, ren2023outofdistributiondetectionselectivegeneration, vazhentsev-etal-2023-efficient, fadeeva-etal-2023-lm} have developed methods to estimate uncertainty in language models, aiming to provide better-calibrated uncertainty estimates or to perform error and out-of-domain (OOD) detection. A relevant approach is utilized in \cite{kadavath2022languagemodelsmostlyknow}, where the authors created a prompt asking the model if the generated prompt is correct. The model's calibration was then measured by the probability of predicting the correct answer when it was correct across the validation set.

One of the most popular directions includes methods that deal with \( p(y|x) \) only. These include softmax maximum probability \cite{hendrycks17baseline}, temperature scaling \cite{guo2017calibration}, and ensembles of deep neural networks for uncertainty estimate \cite{lakshminarayanan2017simple} methods. For auto-regressive models, there are several probabilistic approaches, which utilize the softmax distribution, such as normalized sequence probability \cite{ueffing2007word} and average token-wise entropy \cite{malinin2021uncertaintyestimationautoregressivestructured}. In our work, we follow the recent approach presented in \cite{yang2024towards}, which introduces a maximum entropy estimate for predicting sequence uncertainty. This approach is a better fit for semantic domains where false positive generations must be avoided at all costs. Here, the model’s confidence in sequence prediction is determined by its weakest token prediction.  These methods aim to provide a calibrated estimate that can be utilized later with a threshold. In contrast to uncertainty estimates and subsequent threshold selection, there are methods that incorporate an OOD-detection component in addition to \( p(y|x) \). \citet{chen2023error} developed an additional model to the Text-to-SQL component -- a binary classification model with input of question and generated SQL.

\paragraph{Text-to-SQL}
Text-to-SQL, as a sub-domain of semantic parsing, is deeply influenced by distribution shifts \cite{suhr-etal-2020-exploring, finegan-dollak-etal-2018-improving}. On one side, there is domain shift, where a model trained on one set of databases must perform well on another set. Multiple datasets like SPIDER \cite{yu2018spider} and BIRD \cite{li2024can} are designed to evaluate this aspect. On the other side, there is compositional shift, involving novel SQL query structures in the test dataset. \cite{shaw-etal-2021-compositional, finegan-dollak-etal-2018-improving} explored the models' ability to generalize to novel SQL templates and noted a significant accuracy drop in results. However, these datasets and splits include only answerable questions for the underlying databases. Recently, a ERHSQL benchmark \cite{lee-etal-2024-overview} was presented with a covariate shift, 
featuring unanswerable queries in the test set or those needing external knowledge.

To sum up, while there has been considerable research on uncertainty estimation, such as calibration in semantic parsing \cite{stengel-eskin-van-durme-2023-calibrated} and uncertainty constraints \cite{qin2022sunexploringintrinsicuncertainties} for better calibration, to our knowledge, there is no evident research on selective prediction for probabilistic uncertainty estimation in Text-to-SQL under distribution shifts. In our work, we explore the calibration characteristics of sequence-to-sequence models under different various distribution shift settings (cross-database shift, compositional shift, and covariate shift). Our goal is to detect incorrect generations or generations involving OOD examples, as seen in ERHSQL.

\section{Problem Setup}
We study selective prediction in Text-to-SQL systems under distribution shift settings. Specifically, we examine a Text-to-SQL system consisting of two components: the Text-to-SQL model $\mathcal{Y}$, which takes the natural language utterance $x$ and generates an SQL query $\hat{y}$, and a selective classifier $\mathcal{C}$, which decides whether to output the generated query $\hat{y}$ or abstain based on the uncertainty estimate score $u$. In this section, we formally outline our method for calculating $u$ for generated SQL queries, the selective prediction setup, and the data we evaluated.
In three consecutive studies, we investigate the balance between coverage and accuracy of Text-to-SQL models with a reject option (Sec. \ref{sec:cs1}), model calibration, and the relationship between the confidence of the selective classifier and query characteristics (Sec. \ref{sec:cs2} and \ref{sec:cs3}).

\subsection{Text-to-SQL Models}
In our Text-to-SQL system with reject option, we utilize four models known for their descent ability toward SQL generation. We employ T5-large and T5-3B models from the encoder-decoder family, an in-context-learning GPT-based DAIL-SQL \cite{gao2024text} and a decoder model \llama{} 3, which is fine-tuned using both supervised fine-tuning (SFT) and parameter-efficient fine-tuning (PEFT) techniques with LoRa \cite{hu2022lora}. We fine-tune both the T5 and \llama{} models. To form an input $x$, fine-tuned model receives question $q$ along with database schema $S$. In in-context-learning with DAIL-SQL, we additionally incorporate relevant question-query pairs as examples for ChatGPT prompt. The model is expected to generate a query $\hat{y}$. The hyper-parameters of the fine-tuning are specified in Appendix \ref{app:params}.

\subsection{Uncertainty Estimate}

Given the input sequence $x$ and output sequence $y$ the standard auto-regressive model parameterized by $\theta$ is given by:

$$
P(y|x, \theta) = \prod_{l=0}^{L} P(y_l | y_{<l}, x, \theta) 
$$

Where the distribution of each $y_l$ is conditioned on all previous tokens in a sequence $y_{<l} = {y_0, ..., y_{l-1}}$.

For fine-tuned models, we base our heuristic based on intuition a sequence is only as good as its weakest token prediction $P(y_l | y_{<l}, x, \theta) $ to get the uncertainty estimate $u$ of the whole sequence $y$. If the model soft-max probabilities $p_l$ are close to uniform, the token prediction is less likely to be correct, in contrast to a peak distribution, where the model is certain about token prediction. 

\begin{equation}
\begin{aligned}
    p_l = P(y_l | y_{<l}, x, \theta) \\
    H(p_l) = \sum_{v=0}^{|V|} p_{v} log(p_{v}) \\
    u = max(H_0, ..., H_L) \\
\end{aligned}
\end{equation}

For ChatGPT-based DAIL-SQL, we do not have access to the full vocabulary distribution. Therefore, we utilize the Normalized Sequence Probability modification \cite{ueffing-ney-2005-word}, which was recently featured in one of the EHRSQL shared task solutions \cite{kim-etal-2024-probgate}:
\begin{equation}
    u = \dfrac{1}{|L|}\sum_{l=0}^{|L|} log(p_{l})
\end{equation}

\subsection{Selective Prediction Setting}
In the selective prediction task, given a natural language input $x$, the system outputs $(\hat{y}, u)$ where $\hat{y} \in \mathcal{Y}(x)$ is the SQL query generated by the Text-to-SQL model $\mathcal{Y}$, and $u \in \mathcal{R}$ is the uncertainty estimate. Given a threshold $\gamma \in \mathcal{R}$, the overall Text-to-SQL system predicts the query $\hat{y}$ if $u \geq \gamma$; otherwise, it abstains. The rejection ability is provided by selective classifier $\mathcal{C}$.

Following the experimental setup of \citet{JMLR:v11:el-yaniv10a}, we utilize a testing dataset $D_{tst}$, considered out-of-distribution (OOD) relative to the training dataset $D_{tr}$. We split $D_{tst}$ independently and identically into two data samples: a known OOD sample $D_{known}$ and an unknown OOD sample $D_{unk}$. We use $D_{known}$ to fit our selective classifier or calibrator, and $D_{unk}$ for evaluation.

The main characteristics of the selective classifier $\mathcal{C}$ are its \textit{coverage} and \textit{risk}. Coverage is the fraction of $D_{unk}$ on which the model makes correct predictions, and risk is the error fraction of $D_{unk}$. As the threshold $\gamma$ decreases, both risk and coverage increase. We evaluate our experiments in terms of the risk vs. coverage paradigm. 

To define the target $\hat{y}$ for selective classifiers in Text-to-SQL, we use the inverted execution match metric (EX) for the gold query $g_i$ and predicted query $p_i$ as defined in Equation \ref{eqn:exec_match}. This means that we set the positive class as the presence of an error. 

\begin{equation}
    \hat{y}_{i} =
    \begin{cases}
      0 & \text{if $\text{EX}(g_i) == \text{EX}(p_i)$} \\
      1 & \text{if $\text{EX}(g_i) \neq \text{EX}(p_i)$} \\
    \end{cases}
\end{equation}
\label{eqn:exec_match}

To evaluate results for a particular choice of $\gamma$, we utilize recall and precision metrics.

\begin{figure*}[ht!]
  \centering    
  {\includegraphics[width=0.80\textwidth]{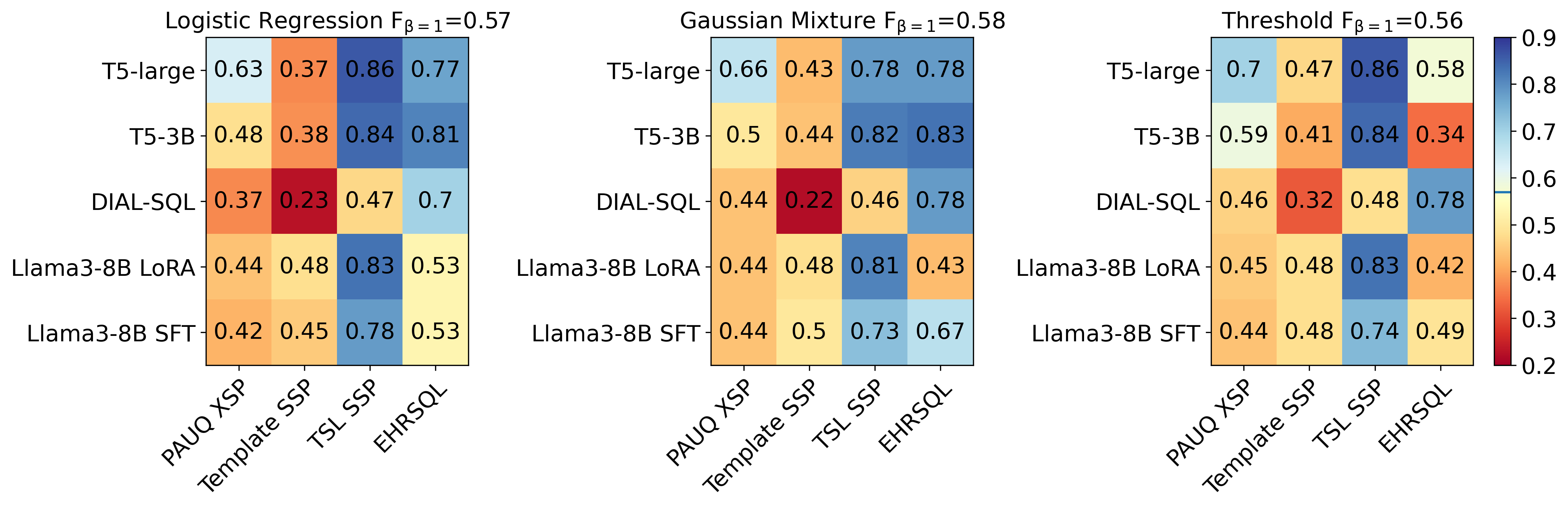}}
  \caption{Heatmaps of $F_{\beta=1}$ per split and model for every selective classifier  (Logistic Regression, Gaussian Mixture, and Threshold).}
  \label{fig:heat_map_f_beta}
\end{figure*}

Coverage refers to our ability to identify and abstain from wrong SQL query generations, while risk corresponds to the proportion of false positive predictions (incorrect queries deemed correct). Recall measures how effectively the system detects errors, and False Discovery Rate (FDR) $(1 - \text{precision})$ indicates the extent to which we abstain from returning correct SQL queries to the user. For a comprehensive assessment of the selective performance across different threshold values $\gamma$, we employ the Area Under the Curve (AUC) metric.

\subsection{Distribution Shift in \task}
We evaluate the uncertainty estimates of \task{} models in distribution shift settings, mimicking various types of shifts: domain shift, compositional shift, and covariate shift. Domain and compositional shifts are full shift examples where $p(x_{tst}) \neq p(x_{tr})$ and $p(y_{tst}|x_{tst}) \neq p(y_{tr}|x_{tr})$, while covariate shift involves only a change in $p(x_{tst}) \neq p(x_{tr})$. Our \task{} pairs $D$ follow $p(x)$ and $p(y|x)$ for training and testing.

To evaluate such distribution shifts, we leverage two \task{} datasets: SPIDER-based PAUQ \cite{bakshandaeva-etal-2022-pauq} and EHRSQL \cite{lee2022ehrsql}. The PAUQ dataset is a refinement of the widely recognized non-synthetic benchmark SPIDER \cite{yu2018spider} for the \task{} task. We prefer English version of PAUQ over SPIDER because it has 8 times fewer empty outputs ($1665 \rightarrow 231$) and 4 times fewer zero-return queries with aggregations (e.g., \textit{maximal}, \textit{minimal}) ($379 \rightarrow 85$). This improvement is crucial, as zero or empty returns can be considered correct when a model generates an executable yet incorrect SQL query, which is undesirable for our study's focus on execution match in the selective classifier. 
EHRSQL is a clinical \task{} dataset that includes pairs of input utterances and expected SQL queries. It covers scenarios where generating an SQL query is not possible for a given question. 

We utulize the following splits to represent different aspects of compositionality \cite{ijcai2020p708,hupkes2023stateoftheart}:

\begin{itemize}
    \item \textbf{PAUQ in cross-database setting} - This setting uses the original SPIDER dataset split, where the data is divided between training and testing sets with no overlap in database structures. During training on $D_{tr}$, the model must learn to generalize to novel database structures found in $D_{tst}$. We refer to this split as \textbf{PAUQ XSP}.
    \item \textbf{PAUQ with template shift in single database setting} - This is the compositional PAUQ split based on templates in a single database setting (same set of databases across $D_{tr}$ and $D_{tst}$), inspired by \cite{finegan-dollak-etal-2018-improving}. This split forces the model to demonstrate its \textit{systematicity} ability—the ability to recombine known SQL syntax elements from $D_{tr}$ to form novel SQL structures in $D_{tst}$. We refer to this split as \textbf{Template SSP}.
    \item \textbf{PAUQ with target length shift in single database setting} - This is another compositional split of the PAUQ dataset, based on the length of SQL queries, in a single database setting \cite{somov-tutubalina-2023-shifted}. Shorter samples are placed in $D_{tr}$, and longer samples in $D_{tst}$, ensuring that all test tokens appear at least once in $D_{tr}$. We refer to this as \textbf{TSL SSP}. It tests the model's \textit{productivity}—its ability to generate SQL queries that are longer than those it was trained on.
    \item \textbf{EHRSQL with unanswerable questions} - This setting uses the original EHRSQL split. Its distinctive feature is the presence of unanswerable questions in $D_{tst}$. These questions cannot be answered using the underlying database content or require external knowledge, making it impossible to generate a correct SQL query. We refer to this split as \textbf{EHRSQL}.
\end{itemize}

For a comparison of our Template SSP and TSL SSP splits with related work, please see Appx. \ref{app:splits}.

\section{Case Study \#1: Selective Text-to-SQL} \label{sec:cs1}

\begin{table*}[t!]
\centering
\begin{tabular}{
>{\columncolor[HTML]{FFFFFF}}r rcccc
>{\columncolor[HTML]{FFFFFF}}c }
\hline
\multicolumn{1}{l}{\cellcolor[HTML]{FFFFFF}{\color[HTML]{333333} }}             & \multicolumn{1}{l}{{\color[HTML]{333333} }}                       & \cellcolor[HTML]{FFFFFF}{\color[HTML]{333333} EX}    & \cellcolor[HTML]{FFFFFF}{\color[HTML]{333333} Recall} & \cellcolor[HTML]{FFFFFF}{\color[HTML]{333333} FDR} & \cellcolor[HTML]{FFFFFF}{\color[HTML]{333333} Result EX} & {\color[HTML]{333333} $\sigma$ Result EX} \\ \hline
\cellcolor[HTML]{FFFFFF}{\color[HTML]{333333} }                                 & \cellcolor[HTML]{FFFFFF}{\color[HTML]{333333} PAUQ XSP}           & \cellcolor[HTML]{D0DABC}{\color[HTML]{333333} 0.634} & \cellcolor[HTML]{9BCDA6}{\color[HTML]{333333} 0.795}        & \cellcolor[HTML]{FFE9A7}{\color[HTML]{333333} 0.218}        & \cellcolor[HTML]{F8D3BD}{\color[HTML]{333333} 0.415}     & {\color[HTML]{333333} 0.097}              \\ \cline{2-7} 
\cellcolor[HTML]{FFFFFF}{\color[HTML]{333333} }                                 & \cellcolor[HTML]{FFFFFF}{\color[HTML]{333333} Template SSP split} & \cellcolor[HTML]{BED6B4}{\color[HTML]{333333} 0.69}  & \cellcolor[HTML]{D4DBBD}{\color[HTML]{333333} 0.624}        & \cellcolor[HTML]{FFA56E}{\color[HTML]{333333} 0.461}        & \cellcolor[HTML]{F0AC9C}{\color[HTML]{333333} 0.229}     & {\color[HTML]{333333} 0.24}               \\ \cline{2-7} 
\cellcolor[HTML]{FFFFFF}{\color[HTML]{333333} }                                 & \cellcolor[HTML]{FFFFFF}{\color[HTML]{333333} TSL SSP split}      & \cellcolor[HTML]{F0AF9E}{\color[HTML]{333333} 0.243} & \cellcolor[HTML]{90CAA2}{\color[HTML]{333333} 0.828}        & \cellcolor[HTML]{FFEBAD}{\color[HTML]{333333} 0.201}        & \cellcolor[HTML]{E7857A}{\color[HTML]{333333} 0.043}     & {\color[HTML]{333333} 0.016}              \\ \cline{2-7} 
\multirow{-4}{*}{\cellcolor[HTML]{FFFFFF}{\color[HTML]{333333} T5-large}}       & \cellcolor[HTML]{FFFFFF}{\color[HTML]{333333} EHRSQL}             & \cellcolor[HTML]{BFD6B4}{\color[HTML]{333333} 0.687} & \cellcolor[HTML]{B9D4B2}{\color[HTML]{333333} 0.705}        & \cellcolor[HTML]{FFFCF2}{\color[HTML]{333333} 0.034}        & \cellcolor[HTML]{CAD9B9}{\color[HTML]{333333} 0.654}     & {\color[HTML]{333333} 0.007}              \\ \hline
\cellcolor[HTML]{FFFFFF}{\color[HTML]{333333} }                                 & \cellcolor[HTML]{FFFFFF}{\color[HTML]{333333} PAUQ XSP}           & \cellcolor[HTML]{B8D4B1}{\color[HTML]{333333} 0.709} & \cellcolor[HTML]{94CBA3}{\color[HTML]{333333} 0.816}        & \cellcolor[HTML]{FFAA72}{\color[HTML]{333333} 0.444}        & \cellcolor[HTML]{F1B3A2}{\color[HTML]{333333} 0.265}     & {\color[HTML]{333333} 0.157}              \\ \cline{2-7} 
\cellcolor[HTML]{FFFFFF}{\color[HTML]{333333} }                                 & \cellcolor[HTML]{FFFFFF}{\color[HTML]{333333} Template SSP split} & \cellcolor[HTML]{AED2AE}{\color[HTML]{333333} 0.738} & \cellcolor[HTML]{D9DCBF}{\color[HTML]{333333} 0.608}        & \cellcolor[HTML]{FFCC88}{\color[HTML]{333333} 0.335}        & \cellcolor[HTML]{F7D0BB}{\color[HTML]{333333} 0.404}     & {\color[HTML]{333333} 0.228}              \\ \cline{2-7} 
\cellcolor[HTML]{FFFFFF}{\color[HTML]{333333} }                                 & \cellcolor[HTML]{FFFFFF}{\color[HTML]{333333} TSL SSP split}      & \cellcolor[HTML]{F2B7A5}{\color[HTML]{333333} 0.282} & \cellcolor[HTML]{88C89E}{\color[HTML]{333333} 0.852}        & \cellcolor[HTML]{FFEEBA}{\color[HTML]{333333} 0.17}         & \cellcolor[HTML]{EA9387}{\color[HTML]{333333} 0.112}     & {\color[HTML]{333333} 0.021}              \\ \cline{2-7} 
\multirow{-4}{*}{\cellcolor[HTML]{FFFFFF}{\color[HTML]{333333} T5-3B}}          & \cellcolor[HTML]{FFFFFF}{\color[HTML]{333333} EHRSQL}             & \cellcolor[HTML]{B7D4B1}{\color[HTML]{333333} 0.712} & \cellcolor[HTML]{72C295}{\color[HTML]{333333} 0.919}        & \cellcolor[HTML]{FFF7DC}{\color[HTML]{333333} 0.086}        & \cellcolor[HTML]{D3DBBD}{\color[HTML]{333333} 0.626}     & {\color[HTML]{333333} 0.027}              \\ \hline
\cellcolor[HTML]{FFFFFF}{\color[HTML]{333333} }                                 & \cellcolor[HTML]{FFFFFF}{\color[HTML]{333333} PAUQ XSP}           & \cellcolor[HTML]{A5CFAA}{\color[HTML]{333333} 0.765} & \cellcolor[HTML]{5DBD8D}{\color[HTML]{333333} 0.982}        & \cellcolor[HTML]{FF8257}{\color[HTML]{333333} 0.577}        & \cellcolor[HTML]{EEA394}{\color[HTML]{333333} 0.188}     & {\color[HTML]{333333} $\le$ 0.01}         \\ \cline{2-7} 
\cellcolor[HTML]{FFFFFF}{\color[HTML]{333333} }                                 & \cellcolor[HTML]{FFFFFF}{\color[HTML]{333333} Template SSP split} & \cellcolor[HTML]{86C79D}{\color[HTML]{333333} 0.86}  & \cellcolor[HTML]{83C79C}{\color[HTML]{333333} 0.868}        & \cellcolor[HTML]{FF3020}{\color[HTML]{333333} 0.846}        & \cellcolor[HTML]{E67E75}{\color[HTML]{333333} 0.014}     & {\color[HTML]{333333} $\le$ 0.01}         \\ \cline{2-7} 
\cellcolor[HTML]{FFFFFF}{\color[HTML]{333333} }                                 & \cellcolor[HTML]{FFFFFF}{\color[HTML]{333333} TSL SSP split}      & \cellcolor[HTML]{C6D8B8}{\color[HTML]{333333} 0.664} & \cellcolor[HTML]{88C89E}{\color[HTML]{333333} 0.854}        & \cellcolor[HTML]{FF704B}{\color[HTML]{333333} 0.636}        & \cellcolor[HTML]{E78178}{\color[HTML]{333333} 0.028}     & {\color[HTML]{333333} $\le$ 0.01}         \\ \cline{2-7} 
\multirow{-4}{*}{\cellcolor[HTML]{FFFFFF}{\color[HTML]{333333} DIAL-SQL}}       & \cellcolor[HTML]{FFFFFF}{\color[HTML]{333333} EHRSQL}             & \cellcolor[HTML]{EFE2C8}{\color[HTML]{333333} 0.542} & \cellcolor[HTML]{9CCDA6}{\color[HTML]{333333} 0.793}        & \cellcolor[HTML]{FFF5D5}{\color[HTML]{333333} 0.105}        & \cellcolor[HTML]{F9D7C1}{\color[HTML]{333333} 0.437}     & {\color[HTML]{333333} $\le$ 0.01}         \\ \hline
\cellcolor[HTML]{FFFFFF}{\color[HTML]{333333} }                                 & \cellcolor[HTML]{FFFFFF}{\color[HTML]{333333} PAUQ XSP}           & \cellcolor[HTML]{B5D3B0}{\color[HTML]{333333} 0.717} & \cellcolor[HTML]{C8D8B8}{\color[HTML]{333333} 0.66}         & \cellcolor[HTML]{FFC080}{\color[HTML]{333333} 0.373}        & \cellcolor[HTML]{F5C4B0}{\color[HTML]{333333} 0.344}     & {\color[HTML]{333333} 0.068}              \\ \cline{2-7} 
\cellcolor[HTML]{FFFFFF}{\color[HTML]{333333} }                                 & \cellcolor[HTML]{FFFFFF}{\color[HTML]{333333} Template SSP split} & \cellcolor[HTML]{C3D7B6}{\color[HTML]{333333} 0.673} & \cellcolor[HTML]{65BF90}{\color[HTML]{333333} 0.958}        & \cellcolor[HTML]{FF6443}{\color[HTML]{333333} 0.673}        & \cellcolor[HTML]{E67C73}{\color[HTML]{333333} 0}         & {\color[HTML]{333333} $\le$ 0.01}         \\ \cline{2-7} 
\cellcolor[HTML]{FFFFFF}{\color[HTML]{333333} }                                 & \cellcolor[HTML]{FFFFFF}{\color[HTML]{333333} TSL SSP split}      & \cellcolor[HTML]{F3BAA8}{\color[HTML]{333333} 0.296} & \cellcolor[HTML]{6FC194}{\color[HTML]{333333} 0.929}        & \cellcolor[HTML]{FFE196}{\color[HTML]{333333} 0.265}        & \cellcolor[HTML]{E78278}{\color[HTML]{333333} 0.032}     & {\color[HTML]{333333} 0.026}              \\ \cline{2-7} 
\multirow{-4}{*}{\cellcolor[HTML]{FFFFFF}{\color[HTML]{333333} Llama3-8B LoRA}} & \cellcolor[HTML]{FFFFFF}{\color[HTML]{333333} EHRSQL}             & \cellcolor[HTML]{CDD9BA}{\color[HTML]{333333} 0.643} & \cellcolor[HTML]{EDE2C7}{\color[HTML]{333333} 0.546}        & \cellcolor[HTML]{FFB77A}{\color[HTML]{333333} 0.403}        & \cellcolor[HTML]{F0AE9E}{\color[HTML]{333333} 0.239}     & {\color[HTML]{333333} 0.202}              \\ \hline
\cellcolor[HTML]{FFFFFF}{\color[HTML]{333333} }                                 & \cellcolor[HTML]{FFFFFF}{\color[HTML]{333333} PAUQ XSP}           & \cellcolor[HTML]{B0D2AF}{\color[HTML]{333333} 0.731} & \cellcolor[HTML]{A8D0AB}{\color[HTML]{333333} 0.756}        & \cellcolor[HTML]{FFA971}{\color[HTML]{333333} 0.449}        & \cellcolor[HTML]{F2B7A5}{\color[HTML]{333333} 0.282}     & {\color[HTML]{333333} 0.002}              \\ \cline{2-7} 
\cellcolor[HTML]{FFFFFF}{\color[HTML]{333333} }                                 & \cellcolor[HTML]{FFFFFF}{\color[HTML]{333333} Template SSP split} & \cellcolor[HTML]{BBD5B3}{\color[HTML]{333333} 0.697} & \cellcolor[HTML]{76C397}{\color[HTML]{333333} 0.907}        & \cellcolor[HTML]{FF9262}{\color[HTML]{333333} 0.524}        & \cellcolor[HTML]{EDA092}{\color[HTML]{333333} 0.173}     & {\color[HTML]{333333} 0.009}              \\ \cline{2-7} 
\cellcolor[HTML]{FFFFFF}{\color[HTML]{333333} }                                 & \cellcolor[HTML]{FFFFFF}{\color[HTML]{333333} TSL SSP split}      & \cellcolor[HTML]{F5C7B3}{\color[HTML]{333333} 0.361} & \cellcolor[HTML]{7FC69B}{\color[HTML]{333333} 0.88}         & \cellcolor[HTML]{FFC987}{\color[HTML]{333333} 0.342}        & \cellcolor[HTML]{E67F76}{\color[HTML]{333333} 0.019}     & {\color[HTML]{333333} 0.016}              \\ \cline{2-7} 
\multirow{-4}{*}{\cellcolor[HTML]{FFFFFF}{\color[HTML]{333333} Llama3-8B SFT}}  & \cellcolor[HTML]{FFFFFF}{\color[HTML]{333333} EHRSQL}             & \cellcolor[HTML]{CEDABB}{\color[HTML]{333333} 0.64}  & \cellcolor[HTML]{BED6B4}{\color[HTML]{333333} 0.688}        & \cellcolor[HTML]{FFF2CA}{\color[HTML]{333333} 0.131}        & \cellcolor[HTML]{FAE5CC}{\color[HTML]{333333} 0.509}     & {\color[HTML]{333333} 0.042}              \\ \hline
\end{tabular}%
\caption{Error detection table for Gaussian Mixture.\textbf{ Recall} stands for accuracy of error SQL detection, the coverage of our selective classifier. \textbf{False discovery rate (FDR)} is the ratio of incorrectly rejected correct SQL generations. Result EX is the Execution match (EM) minus \textbf{FDR}.}
\label{tab:error_detection_for_gmm}
\end{table*}


In this case study, we aim to address the following research questions:
\textbf{RQ1}: Among selective classifiers, which classifier offers the best trade-off between coverage and risk? 
\textbf{RQ2}: What is the impact on the performance of a Text-to-SQL system when the system is expanded with a selective classifier? 
\textbf{RQ3}: What distribution shifts present the most significant challenge for Text-to-SQL with a reject option? 
\textbf{RQ4:}  Given the existence of unanswerable questions in the test set, what types of errors are we more likely to find with a selective classifier?

In our selective classifier methods, we utilize approaches outlined in \cite{JMLR:v11:el-yaniv10a, lee2022ehrsql}, including the threshold-based approach \cite{lee2024trustsqlbenchmarkingtexttosqlreliability}, Logistic Regression, and Gaussian Mixture clustering.

\paragraph{Logistic regression}
We determine parameters $\theta$ using the sigmoid function. During inference, we predict the probability of $u_i$ corresponding to the error prediction based on the probability score of the sigmoid function with fitted parameters.

\begin{equation}
\centering
\begin{aligned}
    p(y_i|u_i, \theta) = \dfrac{1}{1 + e^{-(\theta_0 + \theta_1 u_i)}} \\
    \hat y_i = [p(y_i|u_i, \theta) > 0.5]
\end{aligned}
\end{equation}
\label{eqn:logistic_regression}

\paragraph{Gaussian mixture clustering}
We consider our uncertainty estimates as a combination of two normal distributions, denoted as $\mathcal{N}_z$. The first distribution, $z_0$, is associated with the uncertainty scores $u_i$ of correct generations, while the second distribution, $z_1$, is linked to error generations. We use the expectation-maximization algorithm (EM) to determine the parameters $\mu_z$, $\sigma_z$, and the mixture weight $\pi_z$ for each distribution $z$. 
During inference, we predict the most likely distribution for a given uncertainty estimate $u_i$ using:

\begin{equation}
    \hat y_i = \arg \max_{z} \left( \pi_z \mathcal{N}(u_i \mid \mu_k, \sigma_k) \right)
\end{equation}

\begin{figure*}[t!]
    \centering
    \begin{subfigure}{0.5\textwidth}
        \centering
        \includegraphics[width=\textwidth]{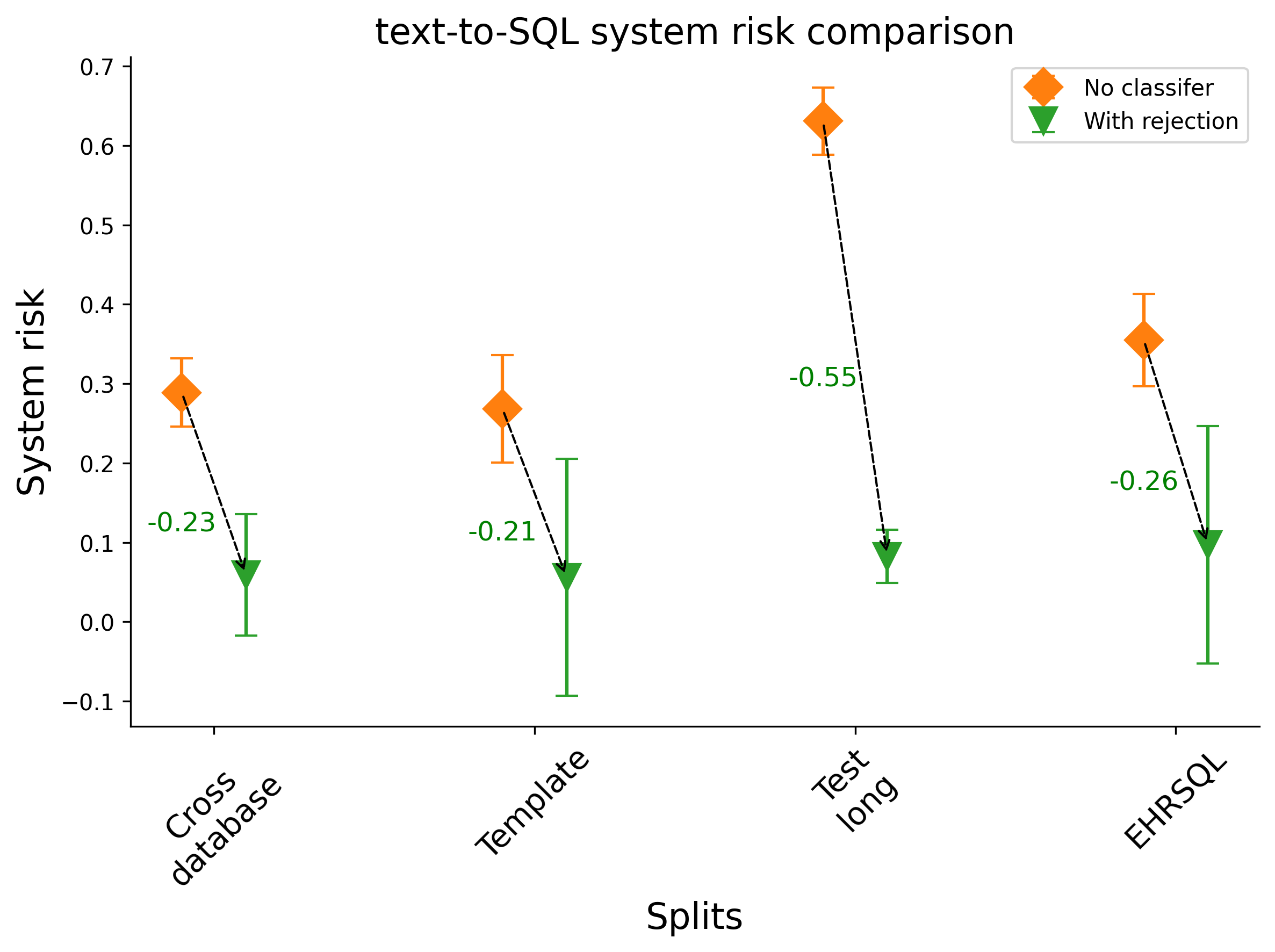}
    \end{subfigure}%
    \begin{subfigure}{0.5\textwidth}
        \centering
        \includegraphics[width=\textwidth]{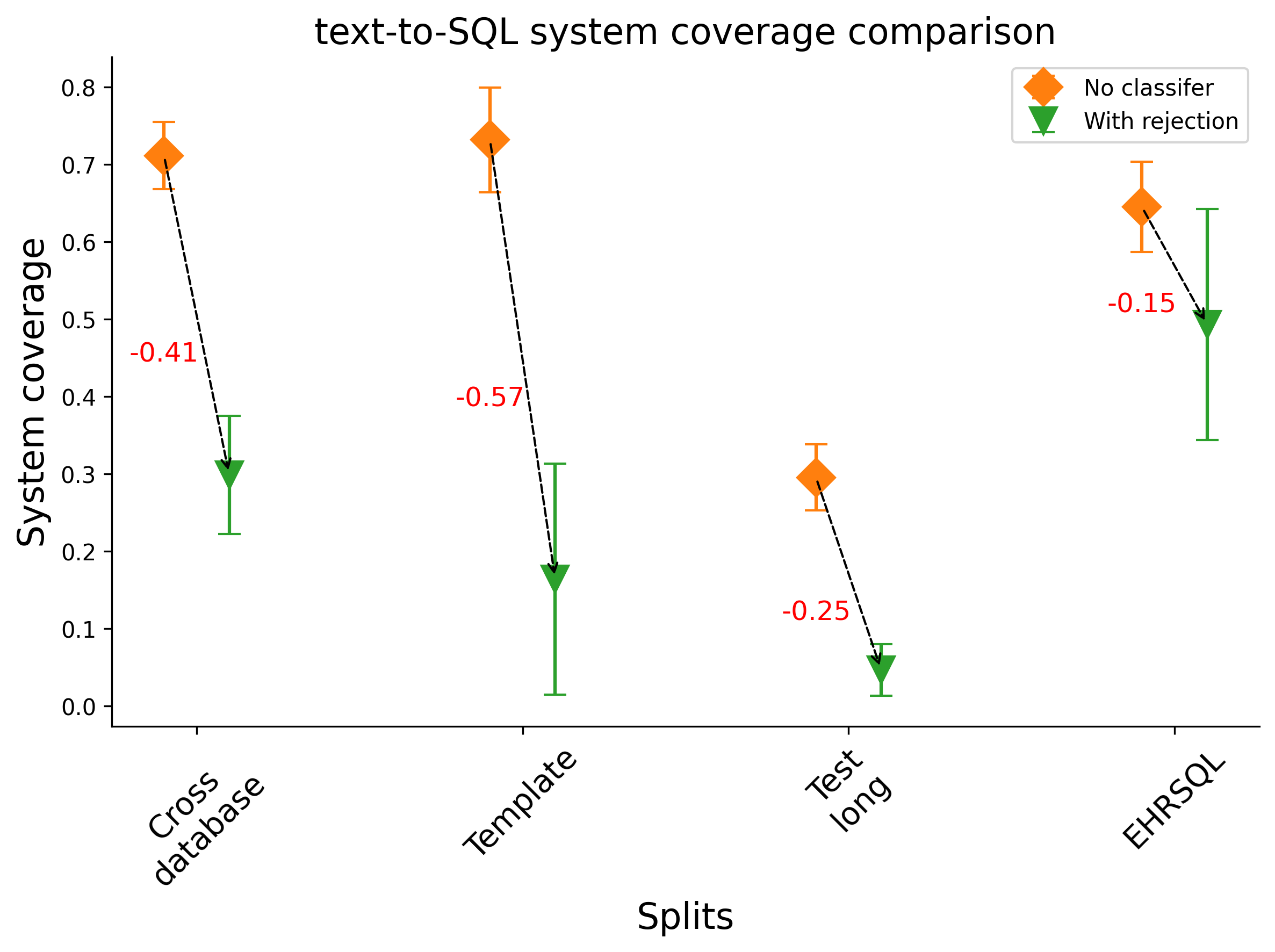}
    \end{subfigure}
    \caption{\textbf{Left:} The system risk decrease with a Gaussian Mixture for every split averaged between all SQL generation models. \textbf{ Right}: The system coverage decrease with the presence of an Gaussian Mixture external classifier for every split averaged between all SQL generation models.}
    \label{fig:risk_and_coverage_comparison}
\end{figure*}

\subsection{Results}
We evaluated our five models on four distinct datasets, using the \(F_{\beta}\) score to compare methods:
$
    F_\beta = ((1 + \beta^2) \text{tp})/((1 + \beta^2) \text{tp} + \text{fp} + \beta^2 \text{fn}),
    \label{eqn:f_beta}
$
tp and fp stand for false and true positives, respectively, fn for false negatives. To address \textbf{RQ1}, we created a heatmap of \(F_{\beta=1}\) scores across all splits and models in Fig. \ref{fig:heat_map_f_beta}. As shown in Fig. \ref{fig:heat_map_f_beta}, the Gaussian Mixture Model demonstrates the best trade-off between precision and recall in a task of error detection. For an in-depth analysis in Appx. \ref{sec:f_beta_scores} we plotted the $F_\beta$ scores for other $\beta$ favoring precision or recall.


In Table \ref{tab:error_detection_for_gmm} (error detection tables for other two methods in Appendix \ref{sec:error_detection_tables}) we are looking for answers related to \textbf{RQ2} and \textbf{RQ3}. This table shows the performance of the Gaussian mixture selective classifier with different models. It focuses on evaluating the coverage and risk assessment of the resulting system. The EHRSQL dataset consistently shows low False Discovery Rate (FDR) values across all model variations, indicating strong error detection capabilities with minimal impact on generated SQL traffic.

Some datasets, such as the TSL SSP split and Template SSP split, present significant challenges for all models when it comes to generating correct SQL queries and rejecting incorrect ones. The Template SSP split poses a particular problem for the selective classifier, exhibiting a high FDR and low TSL SSP initial EX score, indicating limited generalization to longer queries in the testing dataset.

However, we observe that the T5 family models carry lower risk compared to the Chat-GPT solution and \llama{} 3 models, with a lower FDR. This issue with performance is closely linked to the calibration characteristics of the models, which we will further explore in Sec. \ref{sec:cs2}.

Based on selective classification Tab. \ref{tab:error_detection_for_gmm} we built the risk vs coverage comparison in Fig. \ref{fig:risk_and_coverage_comparison} for Gaussian Mixture to answer \textbf{RQ2} and \textbf{RQ3}. As shown in Fig. \ref{fig:risk_and_coverage_comparison} (left), Gaussian Mixture effectively identifies error generations, reducing risk by an average of 80\% across all splits and models, even under distribution shift. However, this comes at the cost of a high false discovery rate (FDR), as also seen in Fig. \ref{fig:risk_and_coverage_comparison} (right). Under these conditions, system coverage remains low, yielding only 1-2 correct generations per 5 requests. However in settings with minimal full shift, such as EHRSQL, selective 
\task{} performs well, achieving an FDR as low as 10\% with some models.

Confirming the results of Fig. \ref{fig:heat_map_f_beta} in Tab. \ref{tab:overall_methods_comparance} we average the scores of Recall, FDR, and Result EX from Appendix D tables, with Gaussian Mixture having the lowest FDR hence does not worsen the Result EX as other two methods.  The error detection tables for Logistic Regression and Threshold-selection are presented in Appx. in Tables \ref{tab:error_detection_for_log_reg} and \ref{tab:error_detection_for_tresh}.

For the further analysis of \textbf{RQ2} and \textbf{RQ3}, we adopt AUC for the selective performance across various threshold values in Appendix \ref{sec:roc_curves} using the probability scores from the Gaussian Mixture classifier. T5-large and T5-3B consistently show superior performance in comparison to other models, especially in the fourth split where they achieve the highest AUC scores (0.93). This suggests that T5-large and T5-3B models are more reliable in terms of Text-to-SQL with a reject option.

To address \textbf{RQ4}, we delved into the types of errors most commonly encountered with the Gaussian Mixture selective classifier in the ERHSQL dataset, as shown in Appendix \ref{sec:ehrsql_bar_plot}. Overall, there is a higher chance of encountering a generation of irrelevant questions as opposed to encountering an incorrect generation. This indicates that all models are fairly confident in generating an incorrect query to a relevant question as opposed to generating a query to an irrelevant one. 
Furthermore, T5-3B, being the most calibrated model as indicated in Table \ref{tab:calibration_methods_comparison}, is capable of accurately detecting even incorrect generations.

\begin{table}
\centering
\begin{tabular}{@{}rccc@{}}
\toprule
\multicolumn{1}{l}{} & Recall   & FDR   & Result EX     \\ \midrule
Gaussian Mixture     & 0.798          & \textbf{0.364} & \textbf{0.251} \\
Logistic Regression  & \textbf{0.873} & 0.469          & 0.145          \\
Threshold            & 0.872          & 0.471          & 0.143          \\ \bottomrule
\end{tabular}%
\caption{Overall methods comparison averaged across all splits and models.}
\label{tab:overall_methods_comparance}
\end{table}

\paragraph{Takeaway 1 (RQ1, RQ2)} The \textit{Gaussian Mixture Model demonstrated the best trade-off between coverage and risk.} The addition of a selective classifier, particularly the \textit{Gaussian Mixture Model, enhances the performance of the Text-to-SQL system by maintaining low False Discovery Rate (FDR) values,} especially on the EHRSQL dataset. This indicates strong error detection capabilities with minimal negative impact on SQL traffic.

\paragraph{Takeaway 2 (RQ3)} \textit{The Template SSP split and TSL SSP split were identified as presenting significant challenges for all models.} These splits demonstrated high FDRs and indicated that the selective classifier struggles with identifying correct SQL queries and effectively rejecting incorrect ones, particularly for longer queries. At the same time, models trained on the EHRSQL dataset under less domain and compositional shifts, show that the error detection method operates much more effectively.

\paragraph{Takeaway 3 (RQ4)}
\textit{Selective classifier has a higher likelihood of spotting generations to irrelevant questions compared to incorrect generations.} This suggests that the models are generally more confident in generating incorrect queries for relevant questions than in generating queries for irrelevant ones. T5-3B, being the most calibrated, effectively detects incorrect generations.






\section{Case Study \#2: Calibration Characteristics} \label{sec:cs2}

\begin{table}[H]
\centering
\begin{tabular}{@{}rccc@{}}
\toprule
\multicolumn{1}{l}{} & MinMax & Platt & Isotonic       \\ \midrule
T5-large             & 0.2    & 0.121  & \textbf{0.117} \\
T5-3B                & 0.17   & 0.108  & \textbf{0.106} \\
Llama3-8B LoRA       & 0.22   & 0.216  & \textbf{0.199} \\
Llama3-8B SFT        & 0.21   & 0.19   & \textbf{0.175} \\
DIAL-SQL             & 0.239  & 0.16   & \textbf{0.152} \\ \bottomrule
\end{tabular}%
\caption{Calibration methods comparison of Brier scores averaged across all splits for each model .}
\label{tab:calibration_methods_comparison}
\end{table}


\begin{figure}[ht]
  \centering    
  {\includegraphics[width=0.45\textwidth]{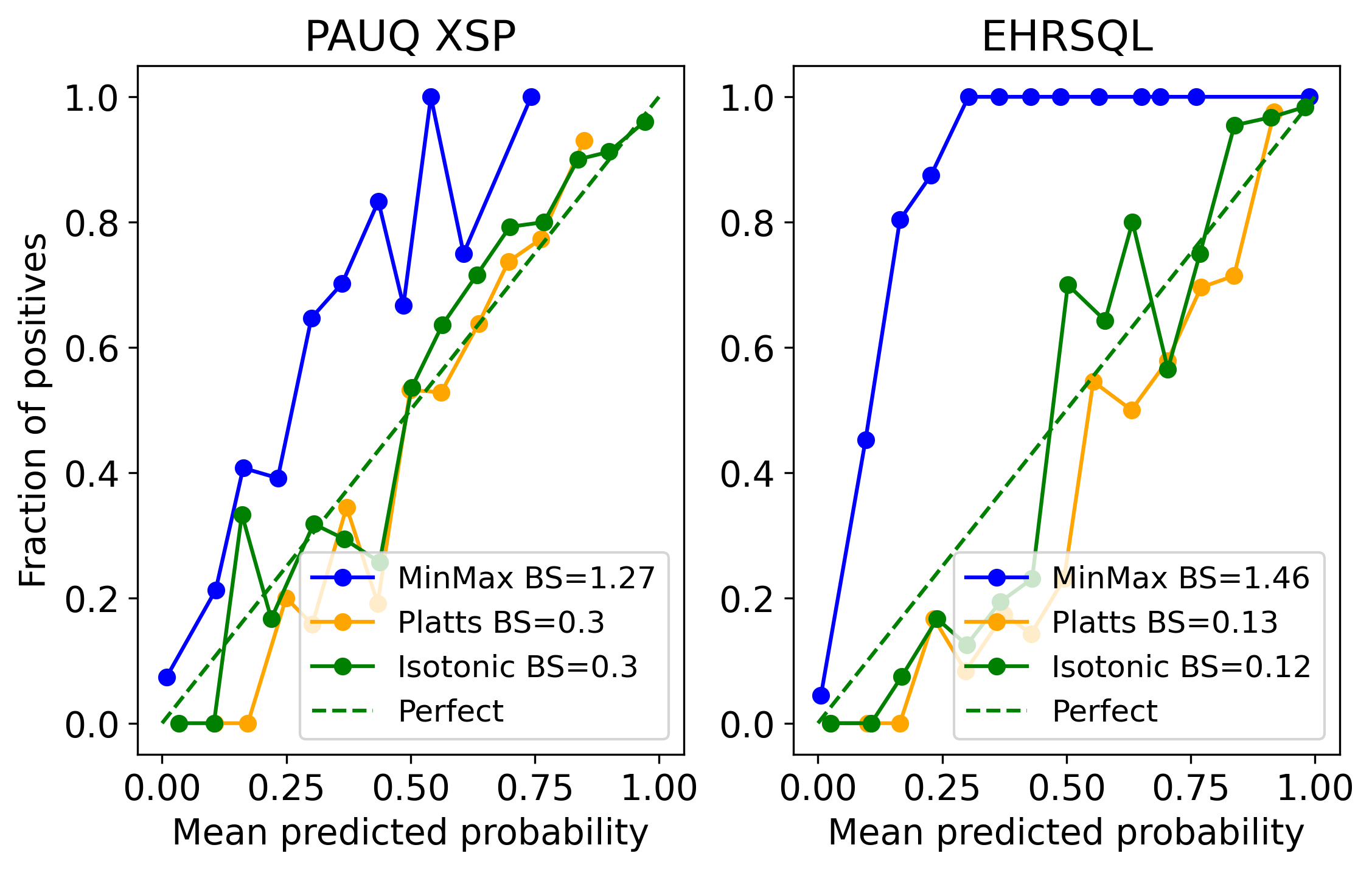}}
  \caption{The calibration effect on T5-3B on PAUQ XSP (cross-database setting) and EHRSQL (single clinical database) compared across MinMax, Platts, and Isotonic calibration (BS stands for Brier score).}
  \label{fig:t5-3b_calibration_methods_comparison}
\end{figure}

\begin{figure*}[ht]
    \centering
    \includegraphics[width=0.95\textwidth]{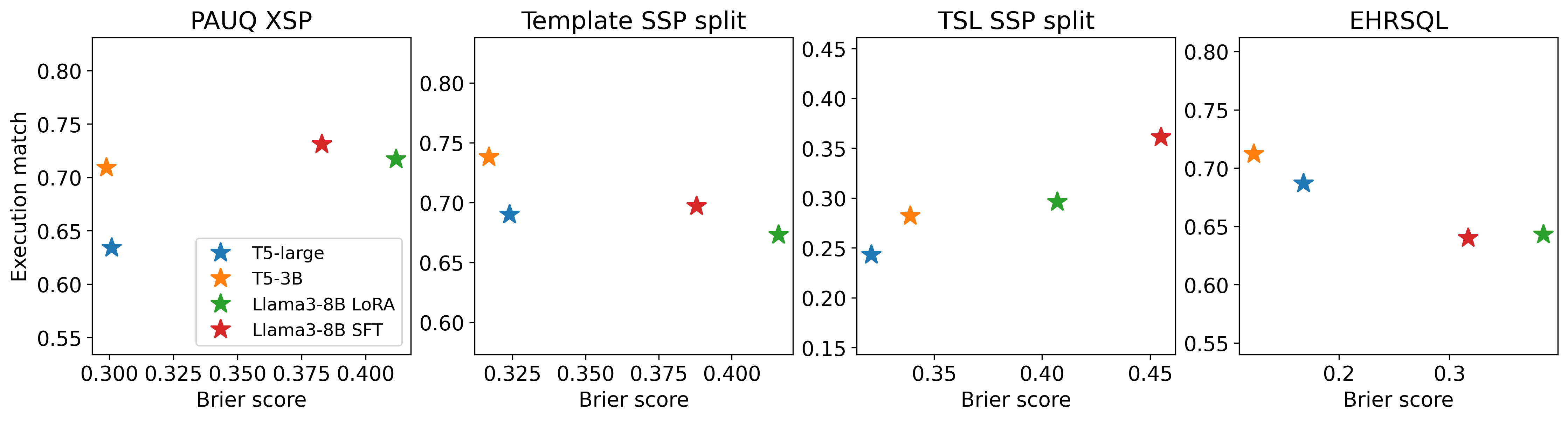}
    \caption{Trade-off plots between execution match and calibration for selected Text-to-SQL models (T5-large, T5-3B, Llama 3 in SFT and LoRa setting).}
    \label{fig:tradeoff}
\end{figure*}

In this section, we will investigate the following research question (\textbf{RQ5}): How do different calibration methods and training datasets influence the calibration of model uncertainty scores, and what trade-offs exist between calibration measures and model execution accuracy? Specifically, the model calibration addresses the question: out of all the instances where we predicted an 80\% chance of a query being correct, how often was the query actually correct? A well-calibrated model would have this proportion close to 80\%. In this case study we want to measure the calibration of the uncertainty estimates.









In contrast to \cite{stengel-eskin-van-durme-2023-calibrated} on calibration in semantic parsing, we measure uncertainty estimates at the sequence level, as this is most relevant for system safety. We define the positive class as an execution match result if $EX(g_i) == EX(p_i)$. 


For calibration of our score \( u \), two calibration methods - Platt calibration and Isotonic calibration - and a naive normalization method (MinMax scaling) were used.

\textbf{MinMax normalization} can be applied here because the maximum entropy estimate is a monotonic function. This allows us to transform the value range from $[0; +\infty]$ to $[0; 1]$. We refer to the calibrated score as $u^c$:
$
    u^c_i = \frac{(u_i - min(u_i))}{max(u_i) - min(u_i)}
$

\textbf{Platt calibration} \cite{platt1999probabilistic} is represented by a logistic regression function from Eq. \ref{eqn:logistic_regression}. The parameters $\theta_0$ and $\theta_1$ are selected on a $D_{known}$ using the maximum likelihood method. 

\textbf{Isotonic regression} \cite{zadrozny2002transforming} involves constructing a piece-wise constant function of the form \( g_m(u_i) = \theta_m \) to transform uncertainty estimates by minimizing the quadratic difference. As \( g_m \) is a piece-wise constant function, the training of this calibration method involves solving the following optimization problem:

\begin{equation}
    \begin{aligned}
            \min_{M,\theta,a} \quad & \sum_{m=1}^M\sum_{i=1}^{N}\mathds{1}(a_m \le u_i < a_{m+1} )(y_i - \theta_m)^2\\
            \textrm{s.t.} \quad & 0 \le a_1 \le a_2 \le ... \le a_{M+1} = 1,\\
                                & \theta_1 \le \theta_2 \le ... \le \theta_M\\
    \end{aligned}
\end{equation}

where \( M \) is the number of function intervals; \( a_1, \ldots, a_{M+1} \) are intervals' boundaries; \( \theta_0, \ldots, \theta_M \) are the values of the function \( g_m \). During fitting on $D_{known}$, Isotonic regression optimizes the heights of the histogram columns for function calibration.


\subsection{Experimental Results}

\paragraph{Evaluation metric} While other calibration comparison methods, such as expected calibration error (ECE), exist, the Brier score \cite{ashukha2020pitfalls} offers a more interpretable means of comparing models. If the model is confident in the positive class (indicated by a high estimate of \( u^c_i \)), then the difference will be minimal.
The Brier scoring method estimate shows the squared difference between the target variable $y_i$ (which can be either 1 or 0) and the predicted probability: 
$
\text{Brier Score} = \sum_{i=1}^{N}(y_i - u^c_i)^2
$

\paragraph{Results} Table \ref{tab:calibration_methods_comparison} presents a comparison of calibration methods using the Brier score, averaged across all data splits. Isotonic regression consistently outperforms both MinMax normalization and Platt calibration across all models, demonstrating its effectiveness in enhancing calibration quality. Notably, as the size of the T5 models increases, the performance of isotonic regression shows improvement. For T5-large, T5-3B, and DIAL-SQL, the advantage of Isotonic calibration over MinMax normalization is significant. In contrast, both Llama3-8B LoRA and SFT exhibit a smaller difference between Isotonic calibration and MinMax, suggesting that the choice of calibration method has less impact on these models. 


Fig. \ref{fig:t5-3b_calibration_methods_comparison} illustrates the calibration curves for Platt, Isotonic, and MinMax for T5 across two datasets. As shown in the figure, the original normalized uncertainty estimate is not calibrated, whereas the calibration methods provide significant improvements.

In Appendix in Fig. \ref{fig:calibration_per_split_full} we present the isotonic calibrations across all models for TSL SSP and EHRSQL splits, averaged over multiple seeds. It is evident that the shifted datasets (PAUQ XSP, Template SSP, and TSL SSP) do not lead to calibrated models. However, in the EHRSQL split with no complete shift, the models demonstrate effective calibration.

Overall in Figure in \ref{fig:tradeoff}, we see that our models exist on a trade-off of calibration and generation quality, with some models being of lower generalization quality but a better calibration. 




\paragraph{Takeaway 4 (RQ5)}

The results indicate that \textit{the original entropy estimate of the models' uncertainty is not calibrated}. \textit{Isotonic regression consistently outperforms other calibration methods like MinMax normalization and Platt calibration} across various models. 
Additionally, \textit{encoder-decoder architecture models are found to be better calibrated compared to decoder-only models}.

\section{Case Study \#3: Query Complexity Analysis} \label{sec:cs3}

In this case study, we investigate the relationship between model confidence in a selective classifier and query complexity, specifically focusing on query length and the number of schema elements in the generated query. Our research question (\textbf{RQ6}): Is the probability of rejection by the selective classifier related to query complexity characteristic?

To address this question, we utilized the Gaussian Mixture Model probabilities (Sec. \ref{sec:cs1}) of the incorrectly generated examples by the T5 model, as T5 models demonstrated the best selective performance across various thresholds (Appendix \ref{sec:roc_curves}).

We assessed query complexity using two key indicators: the length of the generated query(in SQL tokens) and the number of unique schema elements in the generated query. Scatter plots in Fig. \ref{fig:scatter} were constructed for each data split to analyze the relationship between these query characteristics and the confidence of the selective classifier.

Our initial hypothesis was that more complex incorrect queries—characterized by greater length or more schema elements—would correspond to lower probability scores, leading the selective classifier to correctly identify these as incorrect generations. However as the plots show, selective classifier probability does not hold any seeming relation to query complexity.

\paragraph{Takeaway 5 (RQ6)} \textit{Contrary to our hypothesis, we did not observe a proportional decline in model confidence for incorrect queries as query complexity increased. Across all splits, even for the most calibrated models, there was no clear relationship between selective classifier confidence and query complexity.}

\section{Conclusion}
In this paper, we investigated error detection and calibration, utilizing Text-to-SQL LMs with a reject option across general-domain and clinical datasets. 
We believe our findings could enhance the development of more trustworthy Text-to-SQL models in the future.
Future research might concentrate on evaluation across a wider range of datasets and different aspects of compositionality.

\paragraph{Limitations}
Given that our analysis focused on the SPIDER and EHRSQL datasets, the generalizability of our findings may be limited. We concentrated solely on these domains to validate our results, which may not fully capture the variability of noise distributions across different datasets. However, we consider this a minor limitation, as our goal was to observe the models' behavior under distribution shifts rather than to propose and validate a new model with a reject option.

\paragraph{Ethics Statement}
The models and datasets used in this work are publicly available for research purposes.  
All experiments were conducted on four A100 80GB GPUs. Our PyTorch/Hugging Face code will be released with the paper, and we do not anticipate any direct social consequences or ethical issues.

\paragraph{Code}
Code about splitting strategies for measuring compositionality is available at \urlstyle{tt}\url{https://github.com/runnerup96/splitting-strategies}.

The code for training of Text-to-SQL and the selective classification analysis is available at: \urlstyle{tt}\url{https://github.com/runnerup96/error-detection-in-text2sql}.

For execution match calculation we have used   \urlstyle{tt}\url{https://github.com/taoyds/spider} for PAUQ and \urlstyle{tt}\url{https://github.com/glee4810/ehrsql-2024} for EHRSQL.

\section*{Acknowledgments}
We thank Veronica Ganeeva for the figure design. This work was supported by a grant for research centers in the field of artificial intelligence, provided by the Analytical Center for the Government of the Russian Federation in accordance with the subsidy agreement (agreement identifier 000000D730321P5Q0002) and the agreement with the Ivannikov Institute for System Programming of the Russian Academy of Sciences dated November 2, 2021 No. 70-2021-00142.

\bibliography{tacl2021}

\appendix

\section{Models' Hyperparameters}\label{app:params}
In total, similar to \cite{sun2023replication}, training steps towards convergence are $7000$ for PAUQ and $4000$ for EHRSQL for the T5 model. SFT and LoRa were trained for 1 and 3 epochs respectively. Models trained for 3 seed runs -- 1, 42, 123.

\paragraph{T5}
T5-large and T5-3B are tuned with a 0.0001 and 0.00001 learning rate, respectively, with Adafactor \cite{shazeer2018adafactor} with linear decay and batch size of 256.



\paragraph{\llama{} SFT} The batch size is set to 96 and a learning rate of 0.00001. We used the HuggingFace checkpoint available at \urlstyle{tt}\url{https://huggingface.co/meta-llama/Meta-Llama-3-8B}.

\paragraph{\llama{} LoRa} The batch size is set to 96 and a learning rate of 0.0015. $\alpha=16$, dropout$=0.1$, $r=16$, all linear layers are tuned (no bias).

\paragraph{DAIL-SQL} DAIL-SQL\footnote{\urlstyle{tt}\url{https://github.com/BeachWang/DAIL-SQL.git}} with ChatGPT 4 is an in-context learning solution. It embeds the input question with an encoder and finds similar questions based on Euclidean distance. Then, the algorithm appends the top 5 relevant question-query pairs as examples to the ChatGPT prompt (OpenAI API). ChatGPT 4 generates the expected query.

\section{Splits for Compositional Generalization: Comparison with Related Work}\label{app:splits}
Existing evaluations targeting compositional generalization in the Text-to-SQL task include a length split \cite{lake2018generalization}, a template split \cite{finegan2018improving}, a random split, a split based on source length, and a template split on SPIDER \cite{shaw-etal-2021-compositional}. While our template split\footnote{\urlstyle{tt}\url{https://github.com/runnerup96/splitting-strategies}} partially overlaps with existing work, we refine the splitting logic by transforming the target SQL query into a query template. In contrast to  \cite{shaw-etal-2021-compositional,finegan2018improving}, which only masked values and syntax numeric values, our approach masks schema attributes and values using special tokens. To sum up, we apply this masking strategy to our queries and transform SQL queries into templates, which we use for splitting. This modification allows for a more comprehensive evaluation of the models' compositional abilities. For example, a query template could be \verb|SELECT ATTRIBUTE FROM TABLE WHERE| \verb|ATTRIBUTE = NUMERIC|.
Based on the templates, we have prepared target length and template splits. 
These changes have led to this improvement - the template split no longer includes samples with identical structure but different schema elements. 

\section{$F_{\beta}$ Scores Analysis.} \label{sec:f_beta_scores}

\begin{figure}[H]
    \centering
    \begin{subfigure}[b]{0.4\textwidth} 
        \centering
        \includegraphics[width=\textwidth]{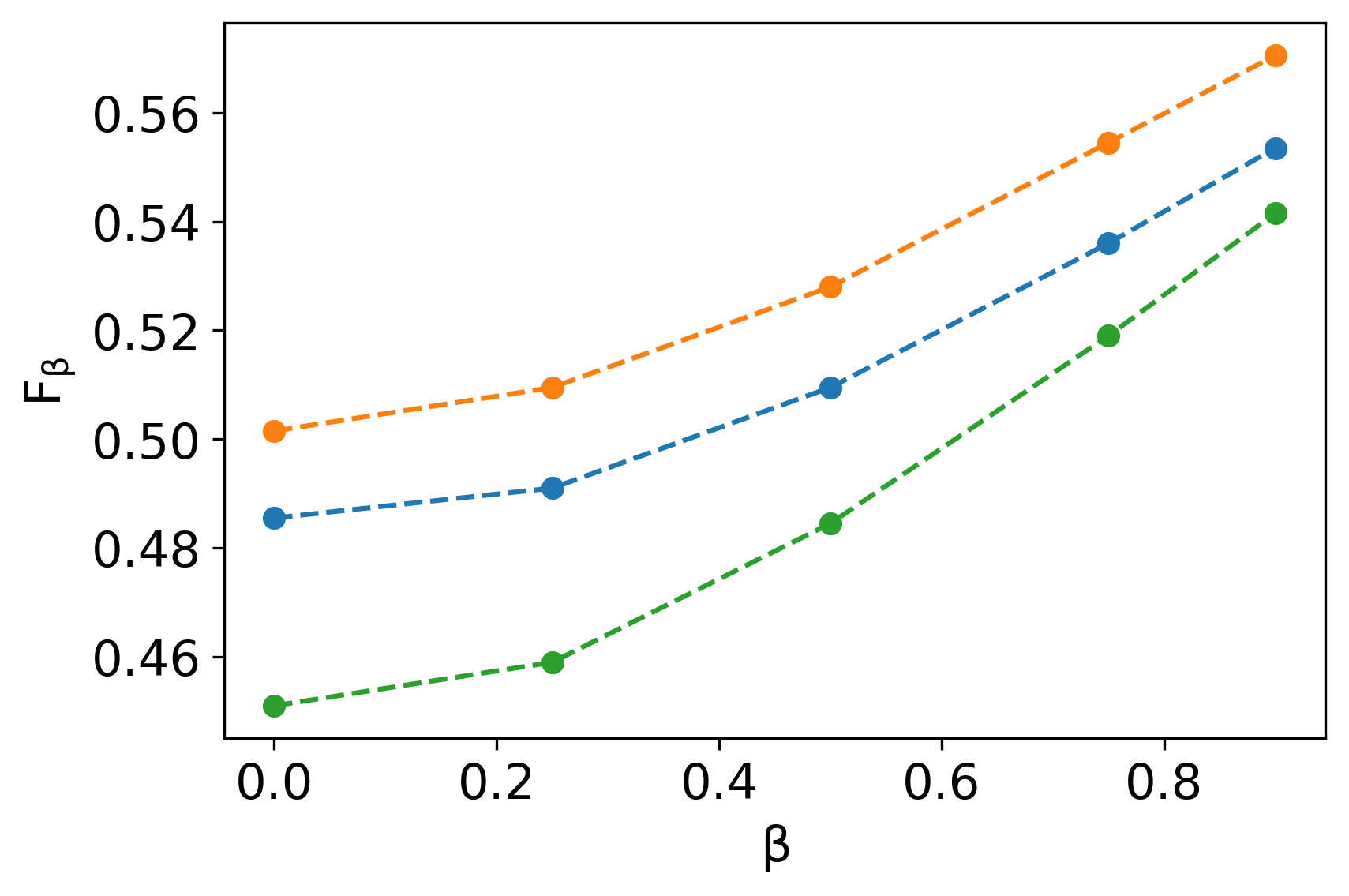}
    \end{subfigure}
    
    \vspace{-0.1cm} 

    \begin{subfigure}[b]{0.4\textwidth}
        \centering
        \includegraphics[width=\textwidth]{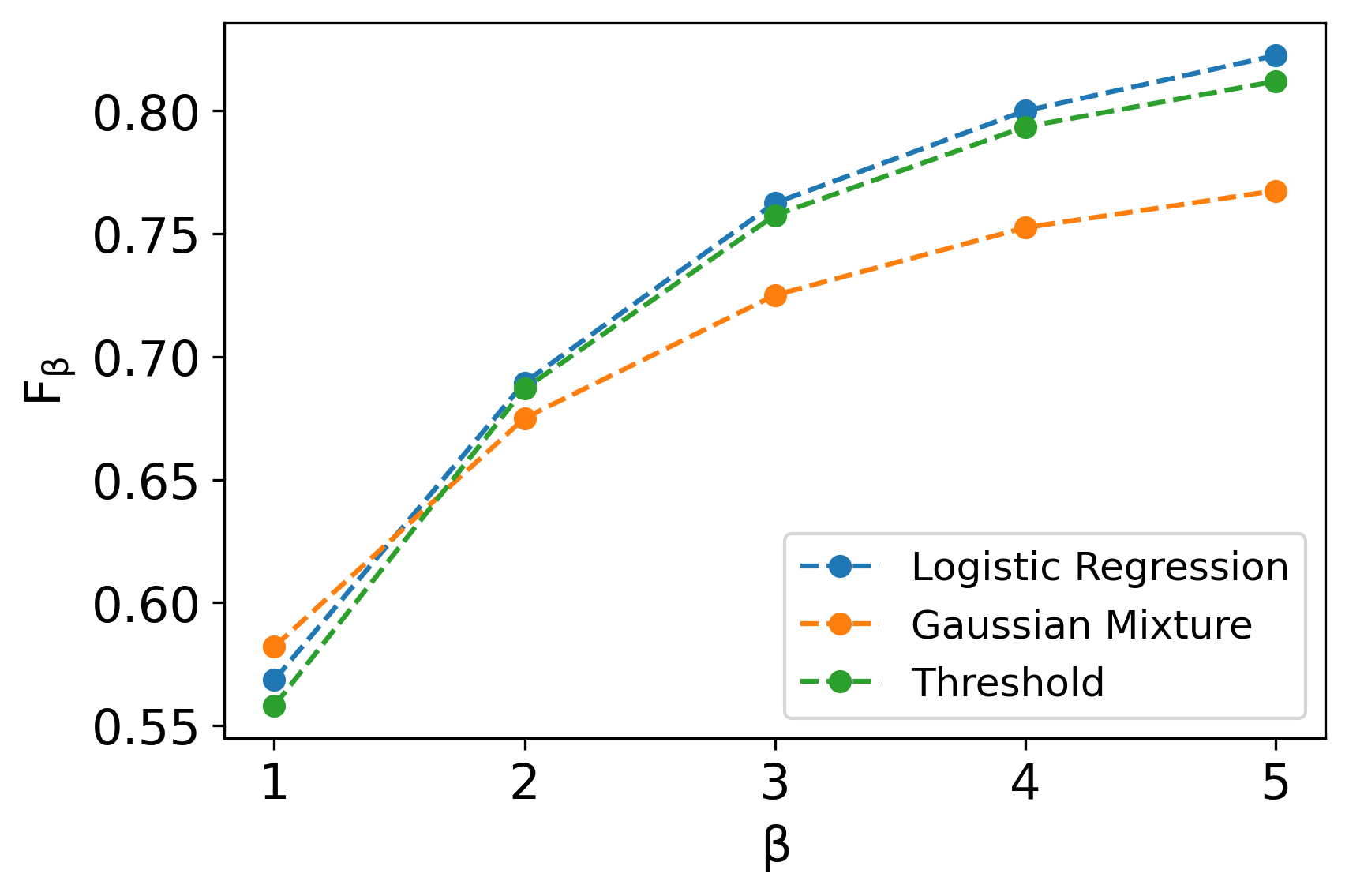}
    \end{subfigure}
    \caption{\textbf{Top:} $F_\beta$ scores for $\beta \in [0;1)$ -- favoring precision over recall. \textbf{Bottom:} $F_\beta$ scores for $\beta \in [1;5]$. -- favoring recall over precision.}
    \label{fig:f_beta_scores}
\end{figure}

\section{Error Detection Tables} \label{sec:error_detection_tables}

Table \ref{tab:error_detection_for_gmm} represents the selective classification metrics for Gaussian Mixture and tables \ref{tab:error_detection_for_log_reg} and \ref{tab:error_detection_for_tresh} for Logistic Regression and Threshold-based selection respectively.

\begin{table*}
\centering
\begin{tabular}{
>{\columncolor[HTML]{FFFFFF}}r rccc
>{\columncolor[HTML]{E67C73}}c 
>{\columncolor[HTML]{FFFFFF}}c }
\hline
{\color[HTML]{333333} }                                                         & \multicolumn{1}{l}{{\color[HTML]{333333} }}                       & \cellcolor[HTML]{FFFFFF}{\color[HTML]{333333} EX}    & \cellcolor[HTML]{FFFFFF}{\color[HTML]{333333} Recall} & \cellcolor[HTML]{FFFFFF}{\color[HTML]{333333} FDR} & \cellcolor[HTML]{FFFFFF}{\color[HTML]{333333} Result EX} & {\color[HTML]{333333} $\sigma$ Result EX} \\ \hline
\cellcolor[HTML]{FFFFFF}{\color[HTML]{333333} }                                 & \cellcolor[HTML]{FFFFFF}{\color[HTML]{333333} PAUQ XSP}           & \cellcolor[HTML]{D0DABC}{\color[HTML]{333333} 0.634} & \cellcolor[HTML]{DBDDC0}{\color[HTML]{333333} 0.601}        & \cellcolor[HTML]{FFF4D2}{\color[HTML]{333333} 0.111}        & \cellcolor[HTML]{F5E4CA}{\color[HTML]{333333} 0.523}     & {\color[HTML]{333333} 0.003}              \\ \cline{2-7} 
\cellcolor[HTML]{FFFFFF}{\color[HTML]{333333} }                                 & \cellcolor[HTML]{FFFFFF}{\color[HTML]{333333} Template SSP split} & \cellcolor[HTML]{BED6B4}{\color[HTML]{333333} 0.69}  & \cellcolor[HTML]{BBD5B3}{\color[HTML]{333333} 0.698}        & \cellcolor[HTML]{FF6D49}{\color[HTML]{333333} 0.644}        & \cellcolor[HTML]{E8857B}{\color[HTML]{333333} 0.046}     & {\color[HTML]{333333} 0.008}              \\ \cline{2-7} 
\cellcolor[HTML]{FFFFFF}{\color[HTML]{333333} }                                 & \cellcolor[HTML]{FFFFFF}{\color[HTML]{333333} TSL SSP split}      & \cellcolor[HTML]{F0AF9E}{\color[HTML]{333333} 0.243} & \cellcolor[HTML]{57BB8A}{\color[HTML]{333333} 1}            & \cellcolor[HTML]{FFE69C}{\color[HTML]{333333} 0.243}        & {\color[HTML]{333333} 0}                                 & {\color[HTML]{333333} $\le$ 0.01}         \\ \cline{2-7} 
\multirow{-4}{*}{\cellcolor[HTML]{FFFFFF}{\color[HTML]{333333} T5-large}}       & \cellcolor[HTML]{FFFFFF}{\color[HTML]{333333} EHRSQL}             & \cellcolor[HTML]{BFD6B4}{\color[HTML]{333333} 0.687} & \cellcolor[HTML]{BCD5B4}{\color[HTML]{333333} 0.694}        & \cellcolor[HTML]{FFFCF3}{\color[HTML]{333333} 0.03}         & \cellcolor[HTML]{C9D8B8}{\color[HTML]{333333} 0.657}     & {\color[HTML]{333333} 0.005}              \\ \hline
\cellcolor[HTML]{FFFFFF}{\color[HTML]{333333} }                                 & \cellcolor[HTML]{FFFFFF}{\color[HTML]{333333} PAUQ XSP}           & \cellcolor[HTML]{B8D4B1}{\color[HTML]{333333} 0.709} & \cellcolor[HTML]{F5E4CA}{\color[HTML]{333333} 0.523}        & \cellcolor[HTML]{FFE297}{\color[HTML]{333333} 0.262}        & \cellcolor[HTML]{F9DAC3}{\color[HTML]{333333} 0.448}     & {\color[HTML]{333333} 0.259}              \\ \cline{2-7} 
\cellcolor[HTML]{FFFFFF}{\color[HTML]{333333} }                                 & \cellcolor[HTML]{FFFFFF}{\color[HTML]{333333} Template SSP split} & \cellcolor[HTML]{AED2AE}{\color[HTML]{333333} 0.738} & \cellcolor[HTML]{7DC59A}{\color[HTML]{333333} 0.885}        & \cellcolor[HTML]{FF573A}{\color[HTML]{333333} 0.718}        & \cellcolor[HTML]{E68076}{\color[HTML]{333333} 0.02}      & {\color[HTML]{333333} 0.008}              \\ \cline{2-7} 
\cellcolor[HTML]{FFFFFF}{\color[HTML]{333333} }                                 & \cellcolor[HTML]{FFFFFF}{\color[HTML]{333333} TSL SSP split}      & \cellcolor[HTML]{F2B7A5}{\color[HTML]{333333} 0.282} & \cellcolor[HTML]{57BB8A}{\color[HTML]{333333} 1}            & \cellcolor[HTML]{FFDC93}{\color[HTML]{333333} 0.282}        & {\color[HTML]{333333} 0}                                 & {\color[HTML]{333333} $\le$ 0.01}         \\ \cline{2-7} 
\multirow{-4}{*}{\cellcolor[HTML]{FFFFFF}{\color[HTML]{333333} T5-3B}}          & \cellcolor[HTML]{FFFFFF}{\color[HTML]{333333} EHRSQL}             & \cellcolor[HTML]{B7D4B1}{\color[HTML]{333333} 0.712} & \cellcolor[HTML]{B3D3AF}{\color[HTML]{333333} 0.724}        & \cellcolor[HTML]{FFFEF9}{\color[HTML]{333333} 0.017}        & \cellcolor[HTML]{BCD5B3}{\color[HTML]{333333} 0.695}     & {\color[HTML]{333333} 0.004}              \\ \hline
\cellcolor[HTML]{FFFFFF}{\color[HTML]{333333} }                                 & \cellcolor[HTML]{FFFFFF}{\color[HTML]{333333} PAUQ XSP}           & \cellcolor[HTML]{A5CFAA}{\color[HTML]{333333} 0.765} & \cellcolor[HTML]{65BF90}{\color[HTML]{333333} 0.958}        & \cellcolor[HTML]{FF4A32}{\color[HTML]{333333} 0.759}        & \cellcolor[HTML]{E67D74}{\color[HTML]{333333} 0.006}     & {\color[HTML]{333333} $\le$ 0.01}         \\ \cline{2-7} 
\cellcolor[HTML]{FFFFFF}{\color[HTML]{333333} }                                 & \cellcolor[HTML]{FFFFFF}{\color[HTML]{333333} Template SSP split} & \cellcolor[HTML]{86C79D}{\color[HTML]{333333} 0.86}  & \cellcolor[HTML]{6DC193}{\color[HTML]{333333} 0.934}        & \cellcolor[HTML]{FF2C1D}{\color[HTML]{333333} 0.859}        & {\color[HTML]{333333} 0.002}                             & {\color[HTML]{333333} $\le$ 0.01}         \\ \cline{2-7} 
\cellcolor[HTML]{FFFFFF}{\color[HTML]{333333} }                                 & \cellcolor[HTML]{FFFFFF}{\color[HTML]{333333} TSL SSP split}      & \cellcolor[HTML]{C6D8B8}{\color[HTML]{333333} 0.664} & \cellcolor[HTML]{76C397}{\color[HTML]{333333} 0.909}        & \cellcolor[HTML]{FF6C48}{\color[HTML]{333333} 0.648}        & \cellcolor[HTML]{E67F75}{\color[HTML]{333333} 0.015}     & {\color[HTML]{333333} $\le$ 0.01}         \\ \cline{2-7} 
\multirow{-4}{*}{\cellcolor[HTML]{FFFFFF}{\color[HTML]{333333} DIAL-SQL}}       & \cellcolor[HTML]{FFFFFF}{\color[HTML]{333333} EHRSQL}             & \cellcolor[HTML]{EFE2C8}{\color[HTML]{333333} 0.542} & \cellcolor[HTML]{E0DEC2}{\color[HTML]{333333} 0.586}        & \cellcolor[HTML]{FFFBEE}{\color[HTML]{333333} 0.044}        & \cellcolor[HTML]{FBE4CC}{\color[HTML]{333333} 0.498}     & {\color[HTML]{333333} $\le$ 0.01}         \\ \hline
\cellcolor[HTML]{FFFFFF}{\color[HTML]{333333} }                                 & \cellcolor[HTML]{FFFFFF}{\color[HTML]{333333} PAUQ XSP}           & \cellcolor[HTML]{B5D3B0}{\color[HTML]{333333} 0.717} & \cellcolor[HTML]{57BB8A}{\color[HTML]{333333} 1}            & \cellcolor[HTML]{FF573A}{\color[HTML]{333333} 0.717}        & {\color[HTML]{333333} 0}                                 & {\color[HTML]{333333} $\le$ 0.01}         \\ \cline{2-7} 
\cellcolor[HTML]{FFFFFF}{\color[HTML]{333333} }                                 & \cellcolor[HTML]{FFFFFF}{\color[HTML]{333333} Template SSP split} & \cellcolor[HTML]{C3D7B6}{\color[HTML]{333333} 0.673} & \cellcolor[HTML]{65BF90}{\color[HTML]{333333} 0.958}        & \cellcolor[HTML]{FF6443}{\color[HTML]{333333} 0.673}        & {\color[HTML]{333333} 0}                                 & {\color[HTML]{333333} $\le$ 0.01}         \\ \cline{2-7} 
\cellcolor[HTML]{FFFFFF}{\color[HTML]{333333} }                                 & \cellcolor[HTML]{FFFFFF}{\color[HTML]{333333} TSL SSP split}      & \cellcolor[HTML]{F3BAA8}{\color[HTML]{333333} 0.296} & \cellcolor[HTML]{57BB8A}{\color[HTML]{333333} 1}            & \cellcolor[HTML]{FFD790}{\color[HTML]{333333} 0.296}        & {\color[HTML]{333333} 0}                                 & {\color[HTML]{333333} $\le$ 0.01}         \\ \cline{2-7} 
\multirow{-4}{*}{\cellcolor[HTML]{FFFFFF}{\color[HTML]{333333} Llama3-8B LoRA}} & \cellcolor[HTML]{FFFFFF}{\color[HTML]{333333} EHRSQL}             & \cellcolor[HTML]{CDD9BA}{\color[HTML]{333333} 0.643} & \cellcolor[HTML]{57BB8A}{\color[HTML]{333333} 1}            & \cellcolor[HTML]{FF6E49}{\color[HTML]{333333} 0.643}        & {\color[HTML]{333333} 0}                                 & {\color[HTML]{333333} $\le$ 0.01}         \\ \hline
\cellcolor[HTML]{FFFFFF}{\color[HTML]{333333} }                                 & \cellcolor[HTML]{FFFFFF}{\color[HTML]{333333} PAUQ XSP}           & \cellcolor[HTML]{B0D2AF}{\color[HTML]{333333} 0.731} & \cellcolor[HTML]{57BB8A}{\color[HTML]{333333} 1}            & \cellcolor[HTML]{FF5337}{\color[HTML]{333333} 0.731}        & {\color[HTML]{333333} 0}                                 & {\color[HTML]{333333} $\le$ 0.01}         \\ \cline{2-7} 
\cellcolor[HTML]{FFFFFF}{\color[HTML]{333333} }                                 & \cellcolor[HTML]{FFFFFF}{\color[HTML]{333333} Template SSP split} & \cellcolor[HTML]{BBD5B3}{\color[HTML]{333333} 0.697} & \cellcolor[HTML]{67BF91}{\color[HTML]{333333} 0.954}        & \cellcolor[HTML]{FF5D3E}{\color[HTML]{333333} 0.697}        & {\color[HTML]{333333} 0}                                 & {\color[HTML]{333333} $\le$ 0.01}         \\ \cline{2-7} 
\cellcolor[HTML]{FFFFFF}{\color[HTML]{333333} }                                 & \cellcolor[HTML]{FFFFFF}{\color[HTML]{333333} TSL SSP split}      & \cellcolor[HTML]{F5C7B3}{\color[HTML]{333333} 0.361} & \cellcolor[HTML]{57BB8A}{\color[HTML]{333333} 1}            & \cellcolor[HTML]{FFC483}{\color[HTML]{333333} 0.361}        & {\color[HTML]{333333} 0}                                 & {\color[HTML]{333333} $\le$ 0.01}         \\ \cline{2-7} 
\multirow{-4}{*}{\cellcolor[HTML]{FFFFFF}{\color[HTML]{333333} Llama3-8B SFT}}  & \cellcolor[HTML]{FFFFFF}{\color[HTML]{333333} EHRSQL}             & \cellcolor[HTML]{CEDABB}{\color[HTML]{333333} 0.64}  & \cellcolor[HTML]{57BB8A}{\color[HTML]{333333} 1}            & \cellcolor[HTML]{FF6E4A}{\color[HTML]{333333} 0.64}         & {\color[HTML]{333333} 0}                                 & {\color[HTML]{333333} $\le$ 0.01}         \\ \hline
\end{tabular}
\caption{Error detection table for Logistic Regression. \textbf{ Recall} stands for accuracy of error SQL detection, the coverage of our selective classifier. \textbf{False discovery rate (FDR)} is the ratio of incorrectly rejected correct SQL generations. Result EX is the Execution match (EM) minus \textbf{FDR}. }
\label{tab:error_detection_for_log_reg}

\centering
\begin{tabular}{
>{\columncolor[HTML]{FFFFFF}}r rcccc
>{\columncolor[HTML]{FFFFFF}}c }
\hline
\multicolumn{1}{l}{\cellcolor[HTML]{FFFFFF}{\color[HTML]{333333} \textbf{}}}    & \multicolumn{1}{l}{{\color[HTML]{333333} \textbf{}}}              & \cellcolor[HTML]{FFFFFF}{\color[HTML]{333333} EX}    & \cellcolor[HTML]{FFFFFF}{\color[HTML]{333333} Recall} & \cellcolor[HTML]{FFFFFF}{\color[HTML]{333333} FDR} & \cellcolor[HTML]{FFFFFF}{\color[HTML]{333333} Result EX} & {\color[HTML]{333333} $\sigma$ Result EX} \\ \hline
\cellcolor[HTML]{FFFFFF}{\color[HTML]{333333} }                                 & \cellcolor[HTML]{FFFFFF}{\color[HTML]{333333} PAUQ XSP}           & \cellcolor[HTML]{D0DABC}{\color[HTML]{333333} 0.634} & \cellcolor[HTML]{90CAA1}{\color[HTML]{333333} 0.83}         & \cellcolor[HTML]{FFEAAD}{\color[HTML]{333333} 0.203}        & \cellcolor[HTML]{F8D6C0}{\color[HTML]{333333} 0.431}     & {\color[HTML]{333333} 0.026}         \\ \cline{2-7} 
\cellcolor[HTML]{FFFFFF}{\color[HTML]{333333} }                                 & \cellcolor[HTML]{FFFFFF}{\color[HTML]{333333} Template SSP split} & \cellcolor[HTML]{BED6B4}{\color[HTML]{333333} 0.69}  & \cellcolor[HTML]{57BB8A}{\color[HTML]{333333} 1}            & \cellcolor[HTML]{FF5F40}{\color[HTML]{333333} 0.69}         & \cellcolor[HTML]{E67C73}{\color[HTML]{333333} 0}         & {\color[HTML]{333333} $\le$ 0.01}    \\ \cline{2-7} 
\cellcolor[HTML]{FFFFFF}{\color[HTML]{333333} }                                 & \cellcolor[HTML]{FFFFFF}{\color[HTML]{333333} TSL SSP split}      & \cellcolor[HTML]{F0AF9E}{\color[HTML]{333333} 0.243} & \cellcolor[HTML]{57BB8A}{\color[HTML]{333333} 1}            & \cellcolor[HTML]{FFE69C}{\color[HTML]{333333} 0.243}        & \cellcolor[HTML]{E67C73}{\color[HTML]{333333} 0}         & {\color[HTML]{333333} $\le$ 0.01}    \\ \cline{2-7} 
\multirow{-4}{*}{\cellcolor[HTML]{FFFFFF}{\color[HTML]{333333} T5-large}}       & \cellcolor[HTML]{FFFFFF}{\color[HTML]{333333} EHRSQL}             & \cellcolor[HTML]{BFD6B4}{\color[HTML]{333333} 0.687} & \cellcolor[HTML]{DEDEC1}{\color[HTML]{333333} 0.591}        & \cellcolor[HTML]{FFE79F}{\color[HTML]{333333} 0.237}        & \cellcolor[HTML]{F9DAC4}{\color[HTML]{333333} 0.45}      & {\color[HTML]{333333} 0.317}         \\ \hline
\cellcolor[HTML]{FFFFFF}{\color[HTML]{333333} }                                 & \cellcolor[HTML]{FFFFFF}{\color[HTML]{333333} PAUQ XSP}           & \cellcolor[HTML]{B8D4B1}{\color[HTML]{333333} 0.709} & \cellcolor[HTML]{78C497}{\color[HTML]{333333} 0.903}        & \cellcolor[HTML]{FFCB88}{\color[HTML]{333333} 0.338}        & \cellcolor[HTML]{F6C9B5}{\color[HTML]{333333} 0.371}     & {\color[HTML]{333333} 0.031}         \\ \cline{2-7} 
\cellcolor[HTML]{FFFFFF}{\color[HTML]{333333} }                                 & \cellcolor[HTML]{FFFFFF}{\color[HTML]{333333} Template SSP split} & \cellcolor[HTML]{AED2AE}{\color[HTML]{333333} 0.738} & \cellcolor[HTML]{57BB8A}{\color[HTML]{333333} 1}            & \cellcolor[HTML]{FF5036}{\color[HTML]{333333} 0.738}        & \cellcolor[HTML]{E67C73}{\color[HTML]{333333} 0}         & {\color[HTML]{333333} $\le$ 0.01}    \\ \cline{2-7} 
\cellcolor[HTML]{FFFFFF}{\color[HTML]{333333} }                                 & \cellcolor[HTML]{FFFFFF}{\color[HTML]{333333} TSL SSP split}      & \cellcolor[HTML]{F2B7A5}{\color[HTML]{333333} 0.282} & \cellcolor[HTML]{57BB8A}{\color[HTML]{333333} 1}            & \cellcolor[HTML]{FFDC93}{\color[HTML]{333333} 0.282}        & \cellcolor[HTML]{E67C73}{\color[HTML]{333333} 0}         & {\color[HTML]{333333} $\le$ 0.01}    \\ \cline{2-7} 
\multirow{-4}{*}{\cellcolor[HTML]{FFFFFF}{\color[HTML]{333333} T5-3B}}          & \cellcolor[HTML]{FFFFFF}{\color[HTML]{333333} EHRSQL}             & \cellcolor[HTML]{B7D4B1}{\color[HTML]{333333} 0.712} & \cellcolor[HTML]{BBD5B3}{\color[HTML]{333333} 0.699}        & \cellcolor[HTML]{FF593B}{\color[HTML]{333333} 0.711}        & \cellcolor[HTML]{E67C73}{\color[HTML]{333333} 0.001}     & {\color[HTML]{333333} 0.002}         \\ \hline
\cellcolor[HTML]{FFFFFF}{\color[HTML]{333333} }                                 & \cellcolor[HTML]{FFFFFF}{\color[HTML]{333333} PAUQ XSP}           & \cellcolor[HTML]{A5CFAA}{\color[HTML]{333333} 0.765} & \cellcolor[HTML]{7FC69B}{\color[HTML]{333333} 0.88}         & \cellcolor[HTML]{FFA871}{\color[HTML]{333333} 0.45}         & \cellcolor[HTML]{F3BEAB}{\color[HTML]{333333} 0.315}     & {\color[HTML]{333333} $\le$ 0.01}    \\ \cline{2-7} 
\cellcolor[HTML]{FFFFFF}{\color[HTML]{333333} }                                 & \cellcolor[HTML]{FFFFFF}{\color[HTML]{333333} Template SSP split} & \cellcolor[HTML]{86C79D}{\color[HTML]{333333} 0.86}  & \cellcolor[HTML]{B2D3AF}{\color[HTML]{333333} 0.725}        & \cellcolor[HTML]{FFB87B}{\color[HTML]{333333} 0.398}        & \cellcolor[HTML]{FADDC6}{\color[HTML]{333333} 0.462}     & {\color[HTML]{333333} $\le$ 0.01}    \\ \cline{2-7} 
\cellcolor[HTML]{FFFFFF}{\color[HTML]{333333} }                                 & \cellcolor[HTML]{FFFFFF}{\color[HTML]{333333} TSL SSP split}      & \cellcolor[HTML]{C6D8B8}{\color[HTML]{333333} 0.664} & \cellcolor[HTML]{7CC599}{\color[HTML]{333333} 0.89}         & \cellcolor[HTML]{FF754E}{\color[HTML]{333333} 0.619}        & \cellcolor[HTML]{E7857B}{\color[HTML]{333333} 0.045}     & {\color[HTML]{333333} $\le$ 0.01}    \\ \cline{2-7} 
\multirow{-4}{*}{\cellcolor[HTML]{FFFFFF}{\color[HTML]{333333} DIAL-SQL}}       & \cellcolor[HTML]{FFFFFF}{\color[HTML]{333333} EHRSQL}             & \cellcolor[HTML]{EFE2C8}{\color[HTML]{333333} 0.542} & \cellcolor[HTML]{A9D0AB}{\color[HTML]{333333} 0.754}        & \cellcolor[HTML]{FFF7DE}{\color[HTML]{333333} 0.083}        & \cellcolor[HTML]{FADCC5}{\color[HTML]{333333} 0.459}     & {\color[HTML]{333333} $\le$ 0.01}    \\ \hline
\cellcolor[HTML]{FFFFFF}{\color[HTML]{333333} }                                 & \cellcolor[HTML]{FFFFFF}{\color[HTML]{333333} PAUQ XSP}           & \cellcolor[HTML]{B5D3B0}{\color[HTML]{333333} 0.717} & \cellcolor[HTML]{6FC194}{\color[HTML]{333333} 0.929}        & \cellcolor[HTML]{FF764F}{\color[HTML]{333333} 0.615}        & \cellcolor[HTML]{EA9185}{\color[HTML]{333333} 0.103}     & {\color[HTML]{333333} 0.029}         \\ \cline{2-7} 
\cellcolor[HTML]{FFFFFF}{\color[HTML]{333333} }                                 & \cellcolor[HTML]{FFFFFF}{\color[HTML]{333333} Template SSP split} & \cellcolor[HTML]{C3D7B6}{\color[HTML]{333333} 0.673} & \cellcolor[HTML]{6EC194}{\color[HTML]{333333} 0.932}        & \cellcolor[HTML]{FF6E4A}{\color[HTML]{333333} 0.641}        & \cellcolor[HTML]{E78278}{\color[HTML]{333333} 0.032}     & {\color[HTML]{333333} 0.045}         \\ \cline{2-7} 
\cellcolor[HTML]{FFFFFF}{\color[HTML]{333333} }                                 & \cellcolor[HTML]{FFFFFF}{\color[HTML]{333333} TSL SSP split}      & \cellcolor[HTML]{F3BAA8}{\color[HTML]{333333} 0.296} & \cellcolor[HTML]{57BB8A}{\color[HTML]{333333} 1}            & \cellcolor[HTML]{FFD790}{\color[HTML]{333333} 0.296}        & \cellcolor[HTML]{E67C73}{\color[HTML]{333333} 0}         & {\color[HTML]{333333} $\le$ 0.01}    \\ \cline{2-7} 
\multirow{-4}{*}{\cellcolor[HTML]{FFFFFF}{\color[HTML]{333333} Llama3-8B LoRA}} & \cellcolor[HTML]{FFFFFF}{\color[HTML]{333333} EHRSQL}             & \cellcolor[HTML]{CDD9BA}{\color[HTML]{333333} 0.643} & \cellcolor[HTML]{BBD5B3}{\color[HTML]{333333} 0.699}        & \cellcolor[HTML]{FF7F55}{\color[HTML]{333333} 0.585}        & \cellcolor[HTML]{E8877D}{\color[HTML]{333333} 0.057}     & {\color[HTML]{333333} 0.026}         \\ \hline
\cellcolor[HTML]{FFFFFF}{\color[HTML]{333333} }                                 & \cellcolor[HTML]{FFFFFF}{\color[HTML]{333333} PAUQ XSP}           & \cellcolor[HTML]{B0D2AF}{\color[HTML]{333333} 0.731} & \cellcolor[HTML]{A4CFA9}{\color[HTML]{333333} 0.769}        & \cellcolor[HTML]{FFA56F}{\color[HTML]{333333} 0.46}         & \cellcolor[HTML]{F1B4A3}{\color[HTML]{333333} 0.271}     & {\color[HTML]{333333} 0.055}         \\ \cline{2-7} 
\cellcolor[HTML]{FFFFFF}{\color[HTML]{333333} }                                 & \cellcolor[HTML]{FFFFFF}{\color[HTML]{333333} Template SSP split} & \cellcolor[HTML]{BBD5B3}{\color[HTML]{333333} 0.697} & \cellcolor[HTML]{78C498}{\color[HTML]{333333} 0.901}        & \cellcolor[HTML]{FF8459}{\color[HTML]{333333} 0.568}        & \cellcolor[HTML]{EB978A}{\color[HTML]{333333} 0.13}      & {\color[HTML]{333333} 0.183}         \\ \cline{2-7} 
\cellcolor[HTML]{FFFFFF}{\color[HTML]{333333} }                                 & \cellcolor[HTML]{FFFFFF}{\color[HTML]{333333} TSL SSP split}      & \cellcolor[HTML]{F5C7B3}{\color[HTML]{333333} 0.361} & \cellcolor[HTML]{7CC599}{\color[HTML]{333333} 0.89}         & \cellcolor[HTML]{FFD28C}{\color[HTML]{333333} 0.314}        & \cellcolor[HTML]{E8857B}{\color[HTML]{333333} 0.047}     & {\color[HTML]{333333} 0.066}         \\ \cline{2-7} 
\multirow{-4}{*}{\cellcolor[HTML]{FFFFFF}{\color[HTML]{333333} Llama3-8B SFT}}  & \cellcolor[HTML]{FFFFFF}{\color[HTML]{333333} EHRSQL}             & \cellcolor[HTML]{CEDABB}{\color[HTML]{333333} 0.64}  & \cellcolor[HTML]{CAD9B9}{\color[HTML]{333333} 0.652}        & \cellcolor[HTML]{FFAF75}{\color[HTML]{333333} 0.427}        & \cellcolor[HTML]{EFA899}{\color[HTML]{333333} 0.213}     & {\color[HTML]{333333} 0.251}         \\ \hline
\end{tabular}
\caption{Error detection table for Threshold-based selection. \textbf{ Recall} stands for accuracy of error SQL detection, the coverage of our selective classifier. \textbf{False discovery rate (FDR)} is the ratio of incorrectly rejected correct SQL generations. Result EX is the Execution match (EM) minus \textbf{FDR}.}
\label{tab:error_detection_for_tresh}
\end{table*}

\section{ROC Curves} \label{sec:roc_curves}

\begin{figure*}[ht!]
  \centering   
  {\includegraphics[width=0.95\textwidth]{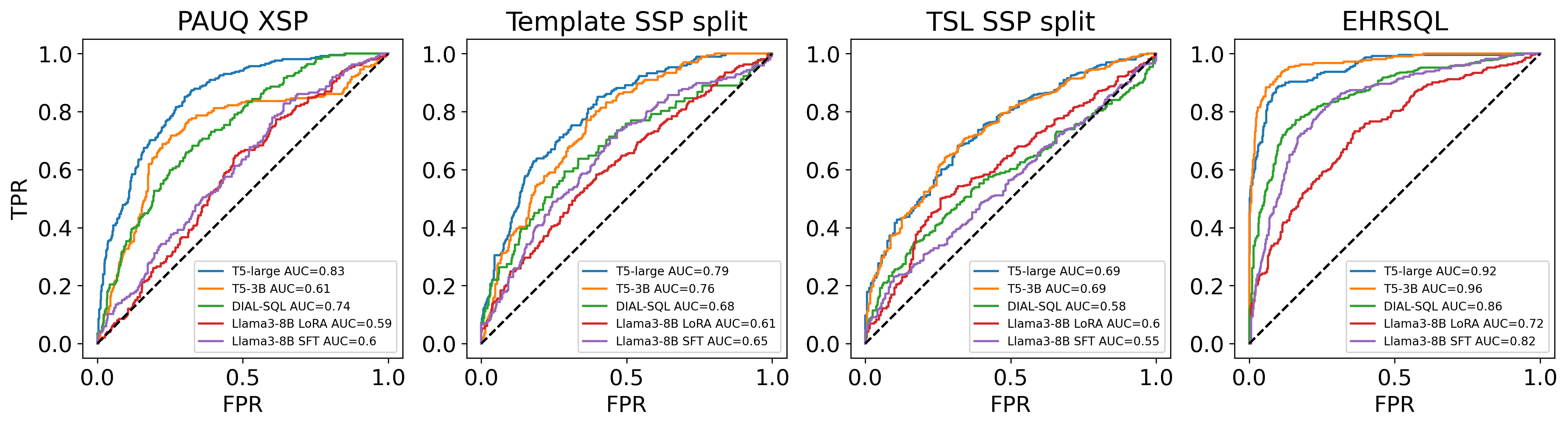}}
  {\includegraphics[width=0.95\textwidth]{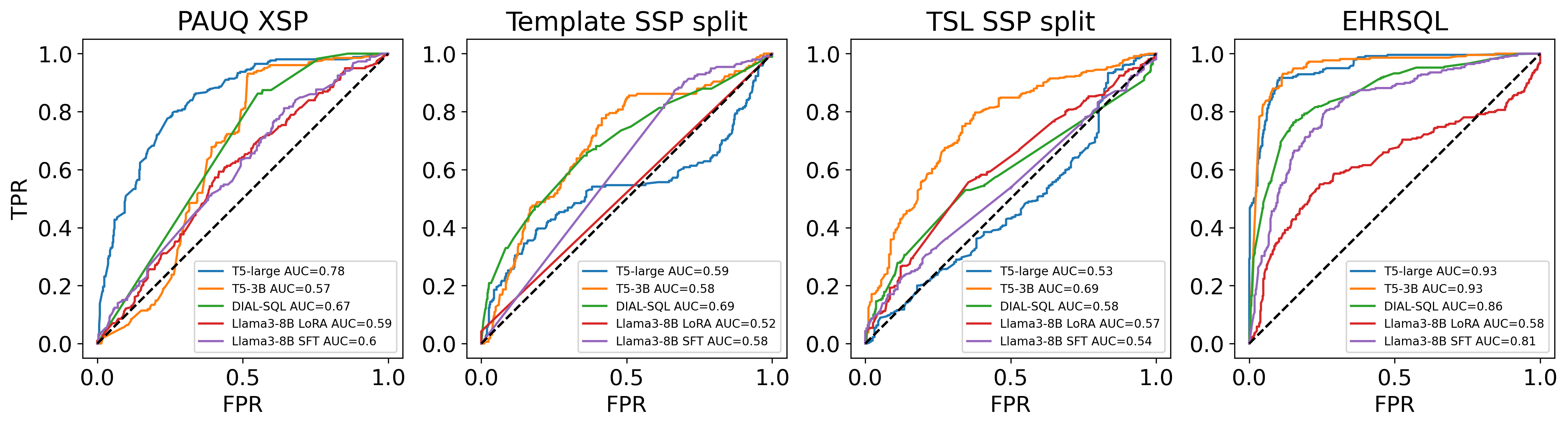}}
  {\includegraphics[width=0.95\textwidth]{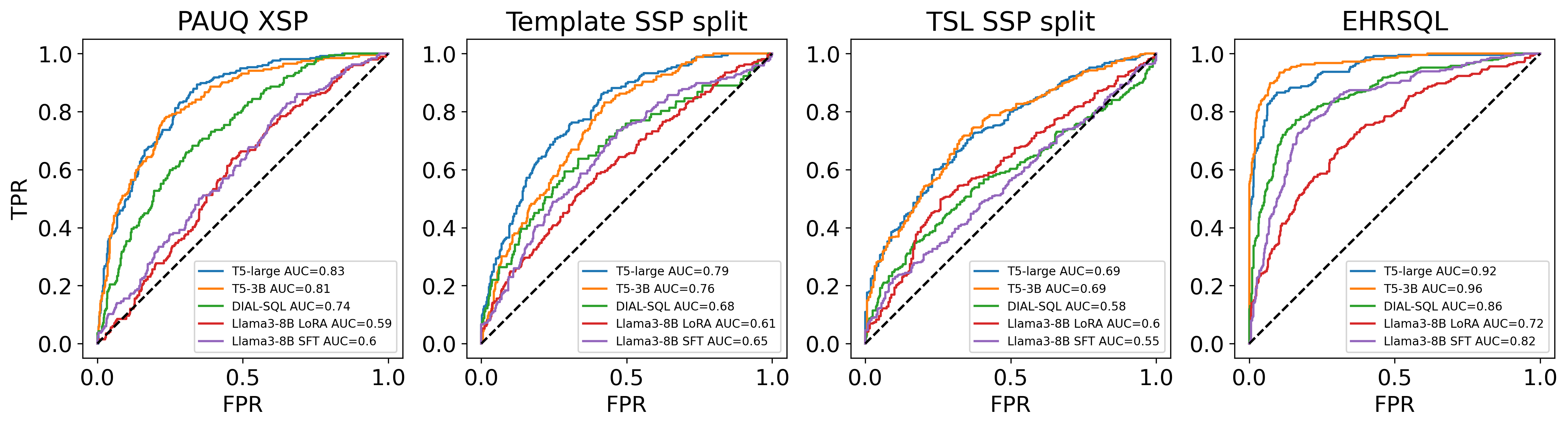}}
  \caption{ROC curves for selected Text-to-SQL models across splits with calculated AUC-ROC score for Logistic Regression (\textbf{Top}), Gaussian Mixture (\textbf{Middle}) and Threshold-based selection (\textbf{Bottom}).}
  \label{fig:roc_curves_for_log_reg_gmm_and_thresh}
\end{figure*}

In Figure \ref{fig:roc_curves_for_log_reg_gmm_and_thresh} we present selective classification ROC curves for detecting an error for Logistic Regression, Gaussian Mixture and Threshold-based selection.

\begin{figure*}
    \centering
    \includegraphics[width=0.9\textwidth]{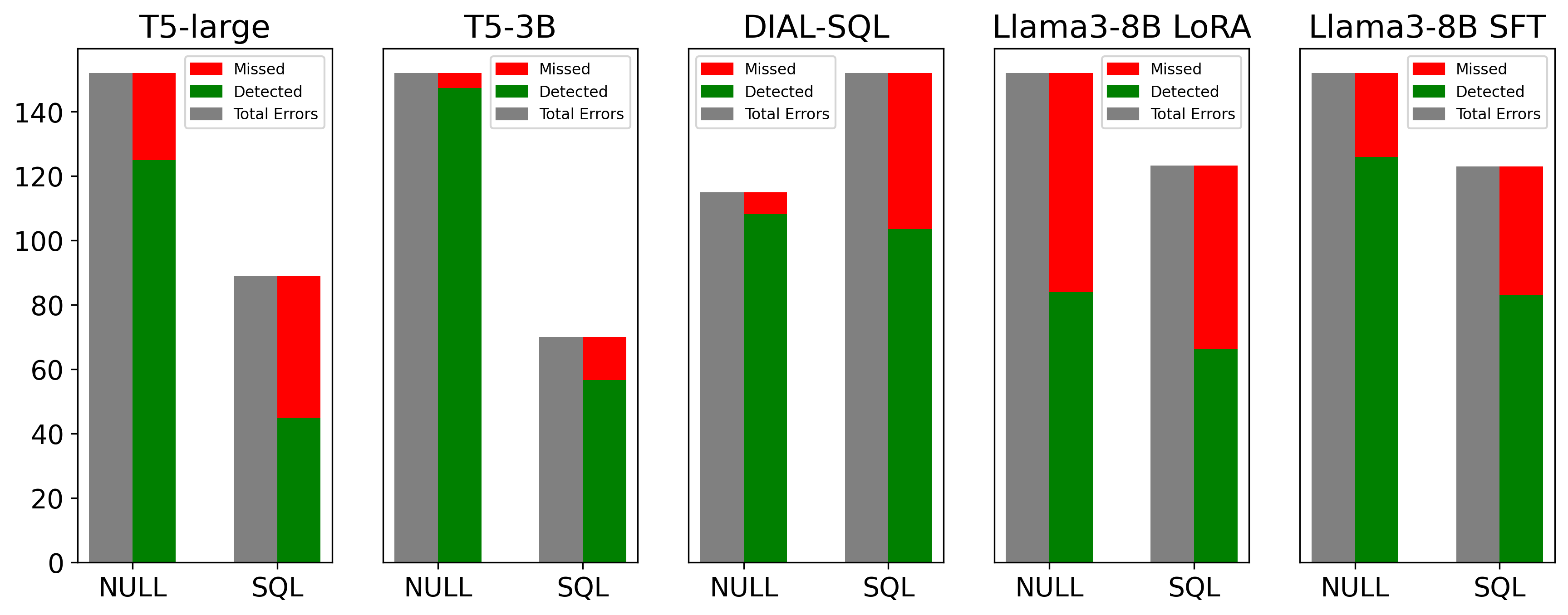}
    \caption{The bar plot for the EHRSQL dataset shows the detection rates of unanswerable queries (NULL bar) and low generalization queries (SQL bar).}
    \label{fig:ehrsql_bar_plot}
\end{figure*}

\section{Error Types on the ERHSQL Dataset} \label{sec:ehrsql_bar_plot}

We performed an analysis of error types most commonly encountered with the Gaussian Mixture selective classifier in the ERHSQL dataset, as shown in Fig. \ref{fig:ehrsql_bar_plot}.

\begin{figure*}[t!]
    \centering
    \includegraphics[width=0.95\textwidth]{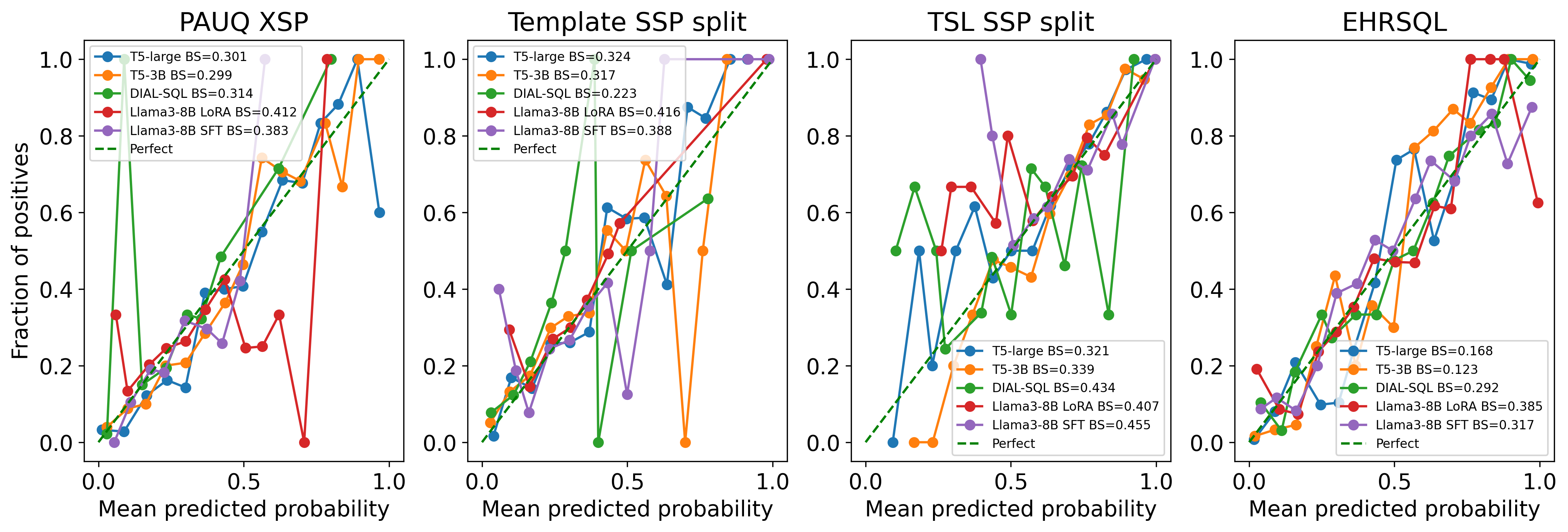}
    \caption{Calibration results using isotonic regression for each model across different splits.}
    \label{fig:calibration_per_split_full}
\end{figure*}

\section{Full Results on Calibration} \label{app:calibration_per_split}

In Fig. \ref{fig:calibration_per_split_full}, we present the isotonic calibrations across all models for TSL SSP and EHRSQL splits, averaged over multiple seeds.

\section{Uncertainty Scores to Query Characteristics Relation} \label{sec:scatter}

\begin{figure*}[ht]
    \centering
    \includegraphics[width=0.95\textwidth]{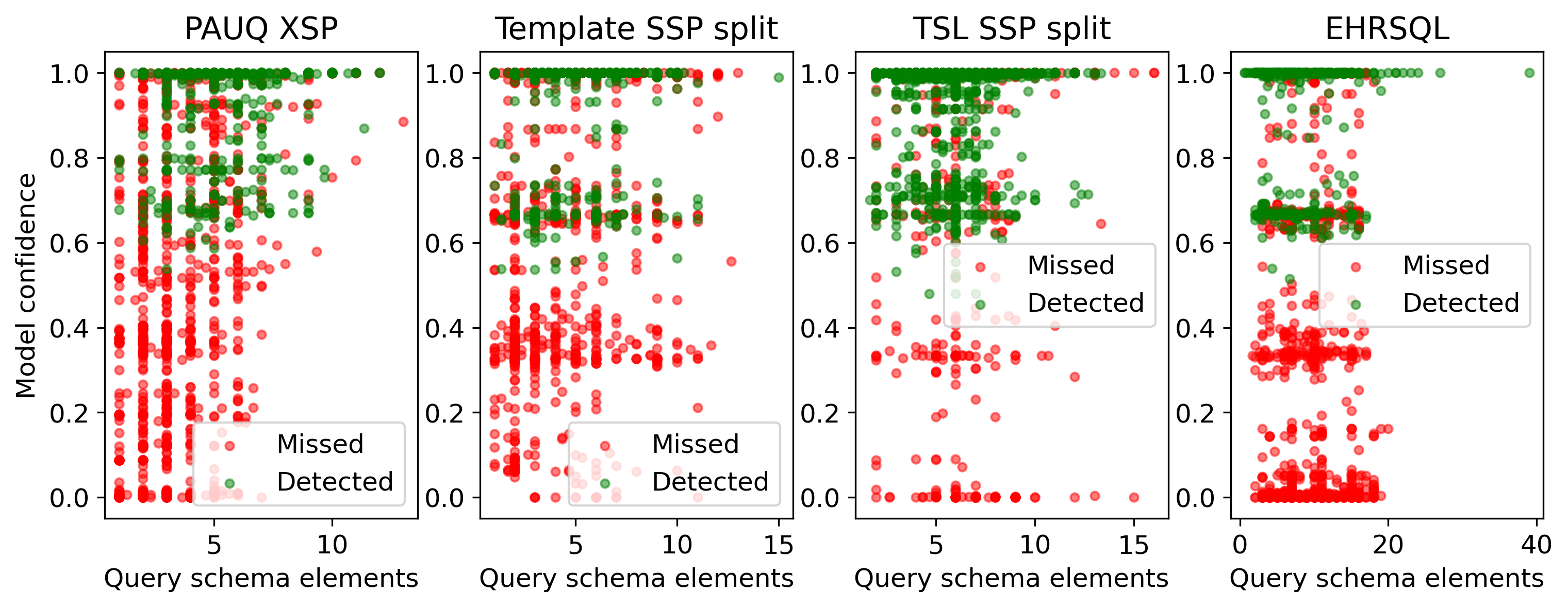}
    \includegraphics[width=0.95\textwidth]{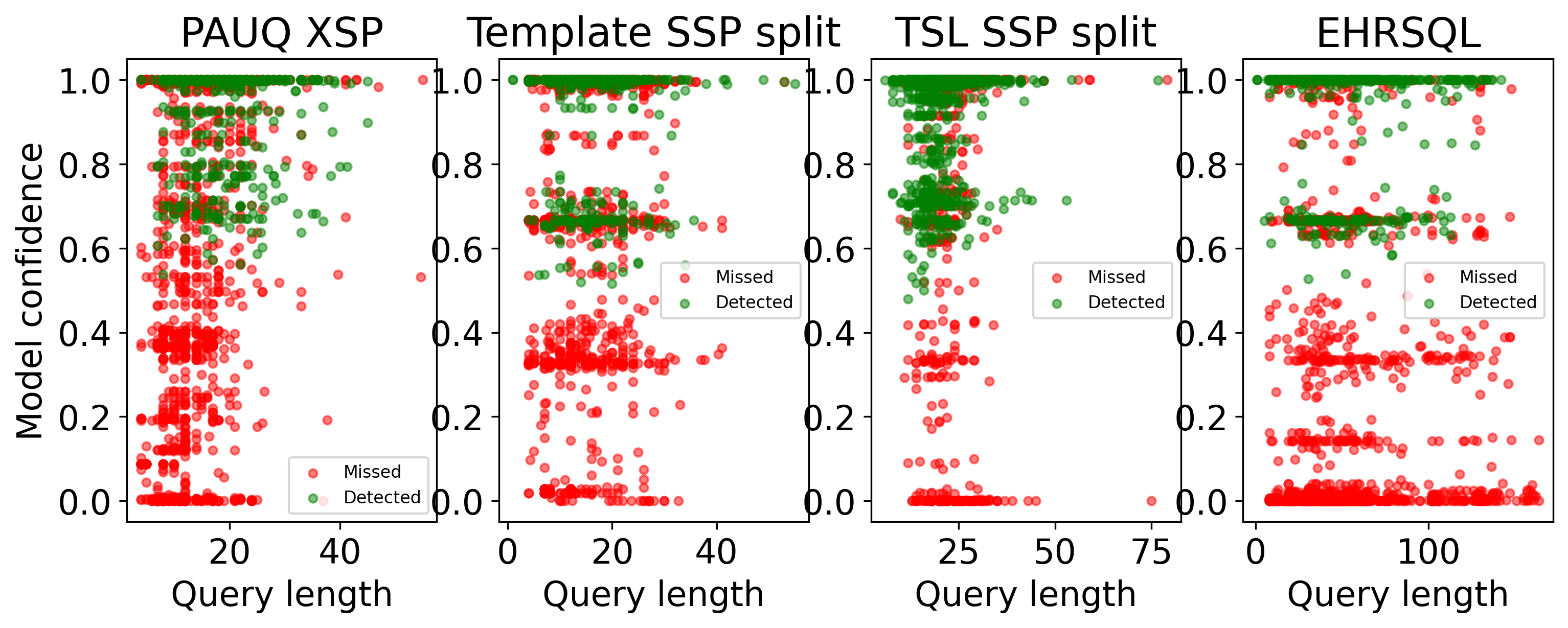}
    \caption{Four scatter plots comparing different splits: PAUQ XSP, Template SSP split, TSL SSP split, and EHRSQL sampled from T5-large and T5-3B \textit{incorrect} predictions. Model confidence is the probability of the sample corresponding to the error cluster in the Gaussian Mixture model. The green points correspond to detected incorrect generations and red points correspond to missed incorrect generations.}
    \label{fig:scatter}
\end{figure*}

Fig. \ref{fig:scatter} includes scatter plots for relation analysis of query characteristics (schema elements in the generated query and query SQL word length) and selective classifier confidence across 4 splits: PAUQ XSP, Template SSP, TSL SSP, and EHRSQL.



\end{document}